\ifdefined\XeTeXversion\else\pdfoutput=1\fi
\documentclass[10pt,twocolumn,letterpaper]{article}

\usepackage[pagenumbers]{wacv}

\newtoggle{inlineappendix}
\toggletrue{inlineappendix}

\usepackage[T1]{fontenc}

\usepackage{amsmath}
\usepackage{amssymb}
\usepackage{amsfonts}
\usepackage{amsthm}
\usepackage{mathtools}
\usepackage{nicefrac}

\usepackage{graphicx}
\usepackage{float}
\usepackage{wrapfig}
\usepackage{booktabs}
\usepackage{longtable}
\usepackage{multirow}
\usepackage{makecell}
\usepackage{adjustbox}

\usepackage{xcolor}
\usepackage{pifont}
\usepackage{enumitem}
\usepackage{microtype}
\usepackage{url}

\usepackage{colortbl}
\definecolor{oursrow}{RGB}{226,236,248}
\newcommand{\newurl}[1]{{\urlstyle{same}\url{#1}}}
\newcommand{\suppref}[1]{\ref*{#1}}

\definecolor{wacvblue}{rgb}{0.21,0.49,0.74}
\usepackage[pagebackref,breaklinks,colorlinks,allcolors=wacvblue]{hyperref}

\def\wacvPaperID{1382}
\def\confName{WACV}
\def\confYear{2027}

\title{PersonaDrive: Human-Style Retrieval-Augmented VLA Agents for\\
Closed-Loop Driving Simulation}

\author{Mahmoud Srewa \enspace Praneetsai Vasu Iddamsetty \enspace Mohammad Abdullah Al Faruque \enspace Salma Elmalaki\\
University of California, Irvine, USA\\
{\tt\small \{msrewa, iddamsep, alfaruqu, salma.elmalaki\}@uci.edu}
}

\begin{document}
\maketitle

\begin{abstract}
\label{sec:abstract}
Closed-loop driving simulators mostly fill their environments with traffic agents that behave alike, whether they are rule-based or learned. Recent methods that control driving style learn it from labels inferred from existing driving recordings, from synthetic data generated to reflect each style, or from reward weights estimated by a large language model (LLM). All of these are indirect estimates of a style, not recordings of people who were instructed to drive in it. We introduce PersonaDrive, a vision-language-action (VLA) driving agent that takes its style from retrieved human demonstrations. The demonstrations come from a new dataset in which $15$ participants drove CARLA routes on a driving simulator under conservative, neutral, and aggressive instructions. PersonaDrive trains a lightweight retrieval head on features from a frozen image encoder and on control signals over per-style databases, and fine-tunes one VLA backbone over the retrieved demonstrations. Switching style only means searching a different FAISS index, with no retraining. On the Bench2Drive benchmark without style conditioning, PersonaDrive improves the Driving Score by $4.6\%$ over SimLingo and by $2.5\%$ over HiP-AD, a gain that fine-tuning without retrieval does not reproduce. With style conditioning, PersonaDrive outperforms the strongest baseline, Drive My Way, in every style, with a Driving Score $6.7\%$ to $11.6\%$ higher. Its driving also follows the instruction, since average speed rises by $19.8\%$ from the conservative to the aggressive style. Each traffic agent is evaluated individually as a driving policy. The dataset and model weights are available on the project page\footnote{\raggedright\newurl{https://pervasiveautonomylab.github.io/personadrive-landpage/}}.
\end{abstract}

\section{Introduction}
\label{sec:introduction}

Autonomous driving research relies heavily on simulation for both training and evaluation. Closed-loop benchmarks such as Bench2Drive~\cite{jia2024bench2drive} are now the standard testbed for end-to-end driving policies. The realism of these simulations depends on perception and physics, and also on how varied the surrounding traffic behaves. Real roads contain a wide range of human driving styles, from conservative through neutral to aggressive, and the interactions between these drivers shape the situations an ego policy must learn to handle. However, current simulators mostly fill the world with rule-based traffic models whose agents behave alike~\cite{lopez2018sumo}. Recent work adds more varied background drivers~\cite{li2026diverse}, but it models this variety with engineered or learned proxies rather than with demonstrations of humans asked to drive in a given style. Adding human driving patterns to simulation and to advanced driver-assistance system (ADAS) pipelines gives much more realistic agent behavior than rule-based or purely learned proxies~\cite{elmalaki2022maconauto, ahadi2021adas}. Moreover, driving behavior differs between people, and even for one driver across situations~\cite{ahadi2021adas, elmalaki2018sentio}. A simulator with a single behavioral mode therefore cannot represent real traffic.

This gap between uniform simulated traffic and varied real traffic limits how well ego policies generalize. It matters most in interactive scenarios, such as merges, unprotected turns, and dense urban driving, where nearby vehicles strongly affect the decisions of the ego vehicle~\cite{mavrogiannis2022b}. Closing the gap requires agents that perceive, reason, and act, as vision-language-action (VLA) models such as SimLingo~\cite{renz2025simlingo} now do, while also covering a controlled range of driving styles.

\begin{figure}[t]
  \centering
  \includegraphics[width=\linewidth]{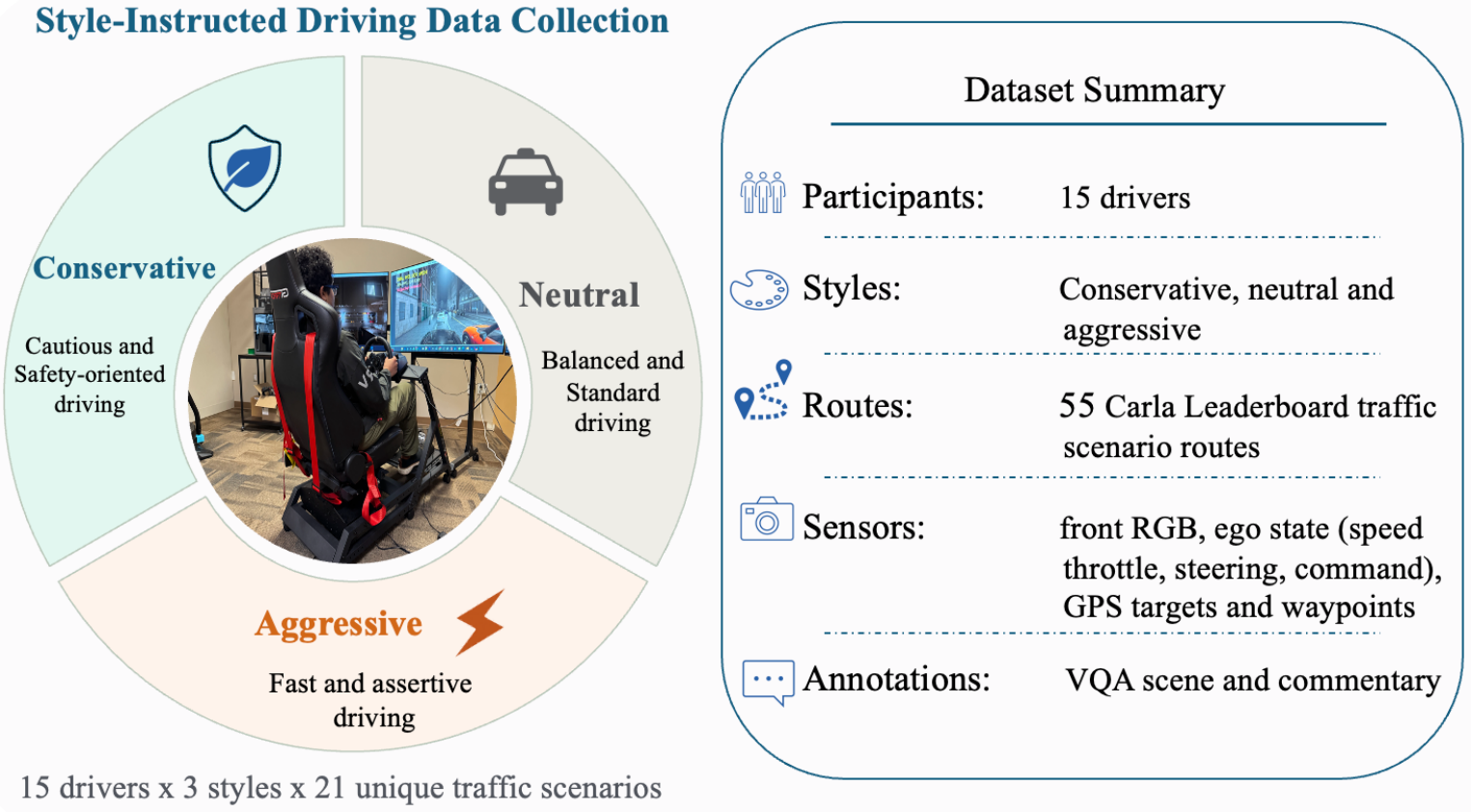}
  \caption{\textbf{Style-instructed driving data collection.} Human participants drive CARLA Leaderboard scenarios on a driving simulator, three times each: once conservatively, once neutrally, and once aggressively. Each drive records the front camera image, the ego state (speed, throttle, steering, command), the GPS targets, and the waypoints. Visual question-answering (VQA) and commentary annotations are added afterwards.}
  \label{fig:data_collection}
\end{figure}

The most closely related work comes from autonomous-vehicle personalization, which seeks to adapt a vehicle's behavior to its occupant's preferences for trust and comfort. StyleDrive~\cite{hao2026styledrive} labels clips of everyday real-world driving as conservative, normal, or aggressive after they were recorded, based on how the vehicle moved. Drive My Way~\cite{wang2026drivemyway} instead asks a large language model (LLM) how much each style should value safety, efficiency, and comfort in a given scenario. Experts review these values, and the driving policy is then trained with reinforcement learning to follow them. Three issues limit the use of these methods for creating a varied population of drivers. First, neither method records how people actually drive when they are told to adopt a style. StyleDrive infers the style from ordinary driving, and Drive My Way learns it from drivers' natural driving and from rewards set by an LLM. Second, retrieval-augmented driving~\cite{yuan2024ragdriver} shows that retrieving past experiences can ground driving decisions. However, it uses one shared database, so it cannot tell which examples fit a particular style. Third, the reward an LLM sets for a style reflects its assumptions about that style, not measurements of humans asked to drive in it, so the learned behavior may differ from real human behavior. An agent that learns directly from human demonstrations does not depend on such assumptions.

We introduce PersonaDrive, a style-conditioned VLA driving agent that addresses these limitations. It produces driving conditioned on different human styles in closed-loop simulation. Our contributions are:
\begin{enumerate}
\item A \textbf{controlled driving-style dataset} (Figure~\ref{fig:data_collection}). Fifteen drivers drove CARLA Leaderboard scenarios in a driver-in-the-loop setup, each of their routes three times: once conservatively, once neutrally, and once aggressively.
Across these three drives, the driver and the route stay the same, and only the style that the driver was asked to follow changes. The dataset has $29{,}777$ frames over $55$ routes and $21$ scenario types. Section~\ref{sec:experiments:setup} describes the data collection.

    \item A \textbf{retrieval-based architecture for driving style}. The style comes from retrieved human demonstrations, and the retrieval is trained on triplets built from the human data. For each frame, called the anchor, a \emph{positive} is a frame with a similar scene that was driven the same way. A \emph{hard negative} has a driving scene similar to the anchor's but was driven differently, for example at another speed or with another command. An \emph{easy negative} differs from the anchor in both scene and action. These triplets teach the retrieval head to find frames that need the same driving, not just frames that look alike. The VLA model then conditions its waypoint prediction on the retrieved examples, so switching style only means searching a different FAISS index~\cite{johnson2017faiss}.

    \item \textbf{Closed-loop results on Bench2Drive}. In closed-loop simulation, the agent's actions change what it sees next, so we test whether it completes routes safely, not only whether it predicts recorded waypoints. We show that the PersonaDrive agent drives better than prior methods, both with and without a style, for example with fewer accidents and traffic violations. We also show that its gain comes from retrieval, not from extra fine-tuning.
\end{enumerate}

In summary, PersonaDrive simulates style-conditioned driving agents that learn from real human drivers. It also improves the driving task itself, finishing routes with fewer accidents than the state-of-the-art VLA model SimLingo~\cite{renz2025simlingo}. Without a style, it raises the Driving Score by $4.6\%$ over SimLingo and by $2.5\%$ over HiP-AD. With a style, it leads in every style while driving faster or slower as instructed. In this work, each PersonaDrive agent is evaluated on its own, as the ego vehicle.

\section{Related Work}
\label{sec:related_work}

\begin{figure*}[t]
  \centering
  \includegraphics[width=\textwidth]{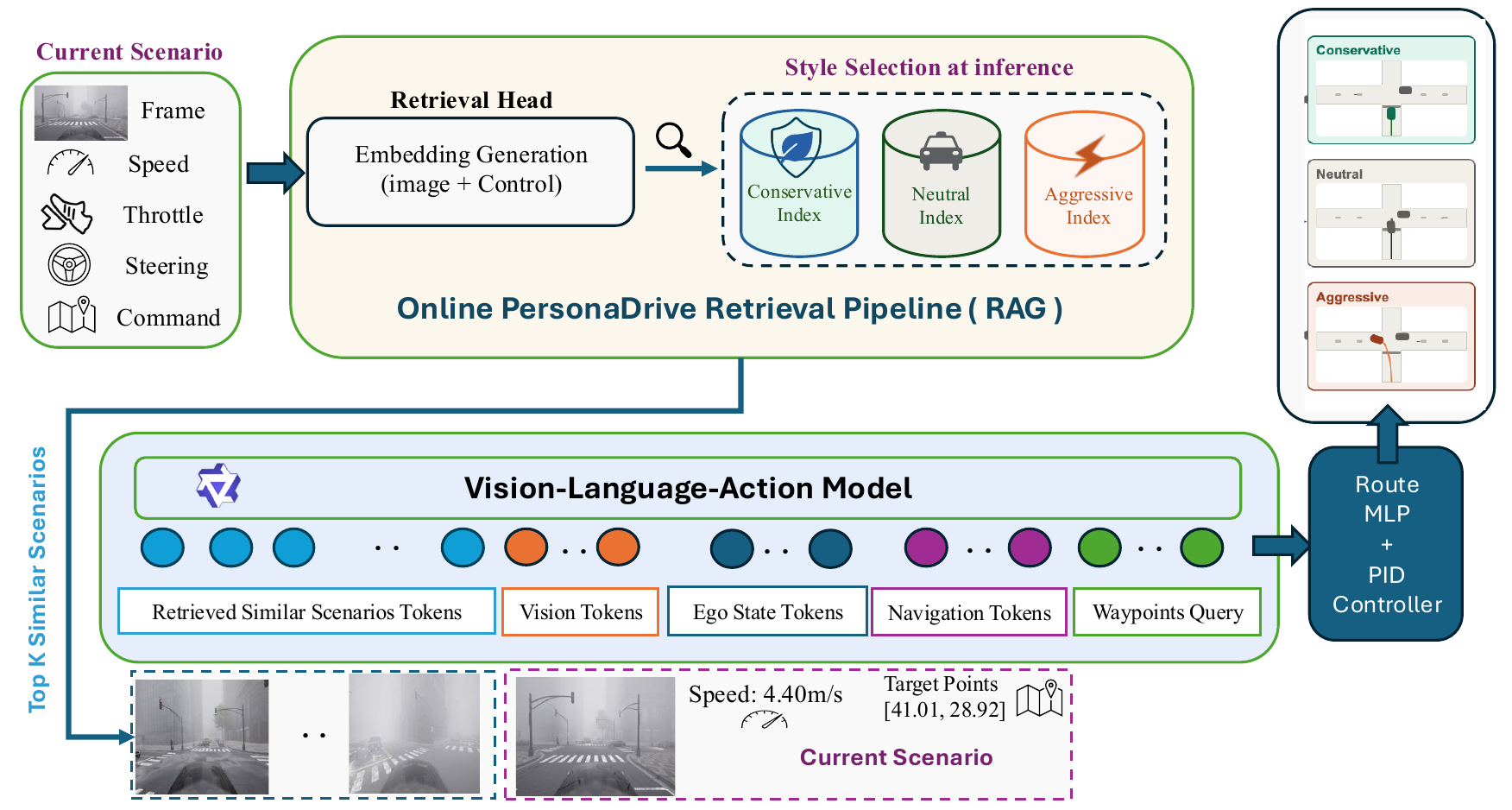}
\caption{\textbf{PersonaDrive framework.} Offline, the style-instructed demonstrations of Figure~\ref{fig:data_collection} fill one search index per style. Online (shown here), the retrieval head encodes the current scene at every step and searches the index of the chosen style. The $K$ most similar scenarios are added to the current observation as in-context tokens for the VLA backbone. The backbone predicts waypoints, which the Route MLP and the PID controllers turn into the style-specific actions shown on the right.}
\label{fig:architecture}
\end{figure*}

\paragraph{Building capable driving agents in simulation.}
Realistic closed-loop simulation depends on non-ego agents that can perceive, reason about, and act on the traffic context. Two threads provide candidate building blocks. First, traffic-agent methods range from rule-based managers (SUMO~\cite{lopez2018sumo}, CARLA's TrafficManager~\cite{dosovitskiy2017carla}) to learned models based on reinforcement learning~\cite{shiroshita2020behaviorally}, autoregressive prediction~\cite{zhou2024behaviorgpt}, and LLM-guided hierarchies~\cite{li2026diverse}. These methods achieve some behavioral variance. However, they define style through engineered parameters or statistical proxies rather than through what humans actually do under a given instruction. Second, end-to-end VLA models such as DriveLM~\cite{sima2024drivelm}, SimLingo~\cite{renz2025simlingo}, FeD~\cite{zhang2024feedback}, AutoVLA~\cite{zhou2025autovla}, and Alpamayo~\cite{wang2025alpamayo} show that a single VLA backbone can serve as a competent ego agent in closed-loop driving. However, each produces only one learned behavioral mode, not a controllable range. PersonaDrive builds on such a VLA backbone and makes its driving style controllable.

\textbf{Style conditioning and retrieval grounding.}
Closer to our setting, recent work conditions driving policies on an explicit style. StyleDrive~\cite{hao2026styledrive} introduces a benchmark and conditions its baselines on one-hot style tokens, which come from labels added after the fact. MAVERIC~\cite{schrum2024maveric} learns a personal style embedding from individual driving data and user questionnaires. Drive My Way (DMW)~\cite{wang2026drivemyway} adapts SimLingo to each style with GRPO~\cite{shao2024deepseekmath}, a reinforcement-learning method, using reward weights that an LLM infers for each scenario and domain experts refine. Driving Style Alignment~\cite{driving_style_alignment} aligns an agent to human demonstrations through feedback from an LLM coach. Other work generates synthetic style data, either by editing expert trajectories to drive faster or slower (D$^3$-MoE~\cite{feng2026d3moe}) or by having a rule-based expert drive at chosen target speeds (Bench2Drive-Speed~\cite{shao2026bench2drivespeed}). All of these methods learn style from proxy signals, such as labels, reward weights, or feedback, or from synthetic datasets. They also target the personalization of one specific user rather than a varied population. On the retrieval side, RAG-Driver~\cite{yuan2024ragdriver} and RealDrive~\cite{ding2025realdrive} ground driving decisions in past experiences. However, both use a single shared database that cannot separate examples by style. RAG-Driver pairs a 7B video language model with a retrieval engine trained with a triplet loss, where positives are chosen by how similar the expert explanations are. It retrieves two examples from one database (BDD-X), writes its explanations and control signals as text, and is evaluated open-loop on recorded data. PersonaDrive differs in both architecture and setting. It mines triplets with a combined image and text similarity, and it builds hard negatives from frames with a different command or speed. It retrieves from separate databases of human demonstrations, one per style. Finally, it feeds the retrieved examples to a VLA whose waypoint heads drive the vehicle in closed-loop simulation, so the retrieved behavior reflects the chosen style. Supp.~Table~\suppref{tab:related_work_comparison} (Supp.~Sec.~\suppref{app:related_work}) compares these frameworks by input, alignment method, and evaluation protocol.

\section{PersonaDrive Framework}
\label{sec:framework}

Figure~\ref{fig:architecture} gives an overview of PersonaDrive. Offline, human participants drive each route once under each style instruction (Figure~\ref{fig:data_collection}, Section~\ref{sec:experiments:setup}, and Supp.~Sec.~\suppref{app:data_collection}). We store the recorded frames in one database per style. Each entry is a \emph{context point} (Section~\ref{sec:framework:context_point}) that records what the driver saw, what they were doing, and what they did next. Online, at every step, a \emph{retrieval head} (Section~\ref{sec:framework:retrieval}) encodes the current scene. It searches the database of the agent's style (conservative, neutral, or aggressive) and returns the $K$ context points with the most similar driving. These points are placed before the current observation in the prompt given to the \emph{VLA backbone} (Section~\ref{sec:framework:sft}). The backbone predicts waypoints, which the Route MLP decodes and two PID controllers turn into steering, throttle, and brake. Training has three stages. Stage~1 mines triplets for the retrieval head (Section~\ref{sec:framework:triplets}). Stage~2 trains the retrieval head. Stage~3 fine-tunes the backbone to read the retrieved points as few-shot examples.

\subsection{Retrieval-Augmented Generation for Style-Aware Driving}
\label{sec:framework:rag}

\textbf{VLA backbone.}
Our starting point is the VLA backbone of SimLingo~\cite{renz2025simlingo}, an InternVL2-1B vision-language model with a dual waypoint head and PID controllers, whose architecture we keep unchanged. It predicts two sets of waypoints in the ego frame. The first is a set of $20$ geometric waypoints $\mathbf{W}^{\text{pos}}_t \in \mathbb{R}^{20 \times 2}$, placed about $1$\,m apart along the path. The second is a set of $10$ temporal waypoints $\mathbf{W}^{\text{vel}}_t \in \mathbb{R}^{10 \times 2}$, placed at $4$\,Hz over the next $2.5$\,s, so their spacing sets the speed. Two MLP heads, the Route MLP in Figure~\ref{fig:architecture}, decode them. Two PID controllers inherited from SimLingo~\cite{renz2025simlingo} then turn $\hat{\mathbf{W}}_t = (\mathbf{W}^{\text{pos}}_t, \mathbf{W}^{\text{vel}}_t)$ into controls, with the geometric waypoints setting the steering and the temporal waypoints setting the throttle and brake. A full description is given in Supp.~Sec.~\suppref{app:model_description}.

\textbf{Motivation.}
The central idea of PersonaDrive is that, at each planning step, the agent looks at how humans drove similar situations under the same style instruction $p \in \mathcal{P} = \{\text{conservative}, \text{neutral}, \text{aggressive}\}$, and uses these past frames as a guide. We do not store style in fixed model weights or in a one-hot token. Instead, we keep the style-$p$ driving history $\mathcal{H}^p$ in its own database and search it at inference time. The retrieved frames come from situations that needed similar reasoning, with a similar road layout, navigational command, and traffic. Their recorded waypoints show how humans drove that situation under instruction $p$. The backbone then conditions its waypoint prediction on them through in-context learning.

\subsection{Stage 1: Frame Sentence Construction and Triplet Generation}
\label{sec:framework:triplets}

Stage~1 creates training data for the retrieval head. It decides which pairs of frames in a style database $\mathcal{H}^p$ show similar driving (positives) and which do not (negatives). Images alone are not enough for this decision, because two frames that look alike may need different actions. We therefore enrich each frame with a natural-language sentence making the full driving context explicit, and use both vision and language similarity to guide mining. Both encoders are frozen. The sentence encoder is used only for mining, while the image encoder is reused by the retrieval head at inference time.

\textbf{Frame sentence construction.}
For each frame $\tau$ in the style history $\mathcal{H}^p$, we build a structured sentence $e^p_\tau$ with seven fields:
\begin{equation}
\small
\label{eq:sentence}
e^p_\tau = \big( v^p_\tau,\ \delta^p_\tau,\ c^p_\tau,\ q^p_\tau,\ r^p_\tau,\ \mathbf{g}^p_\tau,\ \mathbf{W}^{p,\star}_\tau \big),
\end{equation}
where $v^p_\tau$ is the ego speed and $\delta^p_\tau$ is the steering angle. The navigational command $c^p_\tau$ comes from the route planner and is one of \textsc{follow lane}, \textsc{go left}, \textsc{go right}, \textsc{go straight}, \textsc{change lane left}, or \textsc{change lane right}. The annotation $q^p_\tau$ is a visual question-answering (VQA) description of the scene and the traffic~\cite{sima2024drivelm}. The commentary $r^p_\tau$ is free text that explains the action, extended from the commentary of~\cite{renz2025simlingo}, and $\mathbf{g}^p_\tau$ holds the next two GPS target points. The last field, $\mathbf{W}^{p,\star}_\tau \in \mathbb{R}^{N \times 2}$, holds the $N = 10$ waypoints actually driven after frame $\tau$, which are the ego positions in the next $10$ recorded frames, $0.25$\,s apart and covering $2.5$\,s. These waypoints encode the resulting action, both its path and its speed. Together, the VQA annotation and the commentary form the reasoning part of the sentence $e^p_\tau$.

\textbf{Combined similarity and triplet mining.}
Figure~\ref{fig:triplet} shows the Stage~1 mining and the Stage~2 training of the retrieval head. Supp.~Fig.~\suppref{fig:triplet_viewer} shows real examples of mined triplets (anchor, positive, hard negative, easy negative) for all three styles. To measure how similar the driving is in two frames $\tau$ and $\tau'$ of style $p$, we combine two scores. The first is the image similarity from a frozen vision encoder $f_v$ (SigLIP~\cite{zhai2023siglip}), computed on the front-camera images $\mathbf{I}^p_\tau$ and $\mathbf{I}^p_{\tau'}$ of the two frames. The second is the text similarity from a frozen sentence encoder $f_t$ (BGE-M3~\cite{chen2024bgem3}). With $\mathrm{sim}(\cdot,\cdot)$ the cosine similarity, the combined score is
\begin{equation}
\small
\label{eq:combined_sim}
\begin{aligned}
\sigma^p_{\tau,\tau'} \;=\;\;
    & \lambda_{\text{img}} \cdot \mathrm{sim}\!\left( f_v(\mathbf{I}^p_\tau),\, f_v(\mathbf{I}^p_{\tau'}) \right) \\[2pt]
  + \, & \lambda_{\text{txt}} \cdot \mathrm{sim}\!\left( f_t(e^p_\tau),\, f_t(e^p_{\tau'}) \right),
\end{aligned}
\end{equation}
where $\lambda_{\text{img}} + \lambda_{\text{txt}} = 1$. The text term separates frames that look alike but differ in command, speed, or reasoning, such as two similar intersections where the driver turns at one and goes straight at the other. All triplets are mined within a single style database. Within each route, frames are subsampled with a stride of $5$ before mining. This acts as a denoising step, because near-duplicate consecutive frames would otherwise dominate as trivial positives and collapse the contrastive signal (Supp.~Sec.~\suppref{app:frame_selection}).

For each anchor frame $\xi^p_\tau$, which is the context point of Section~\ref{sec:framework:context_point}, we draw three kinds of training samples from the same style database $\mathcal{H}^p$. The thresholds $N^{+}$ and $N^{-}$ are the top and bottom percentiles that bound the positive and negative pools:

\begin{figure*}[t]
  \centering
  \includegraphics[width=\textwidth]{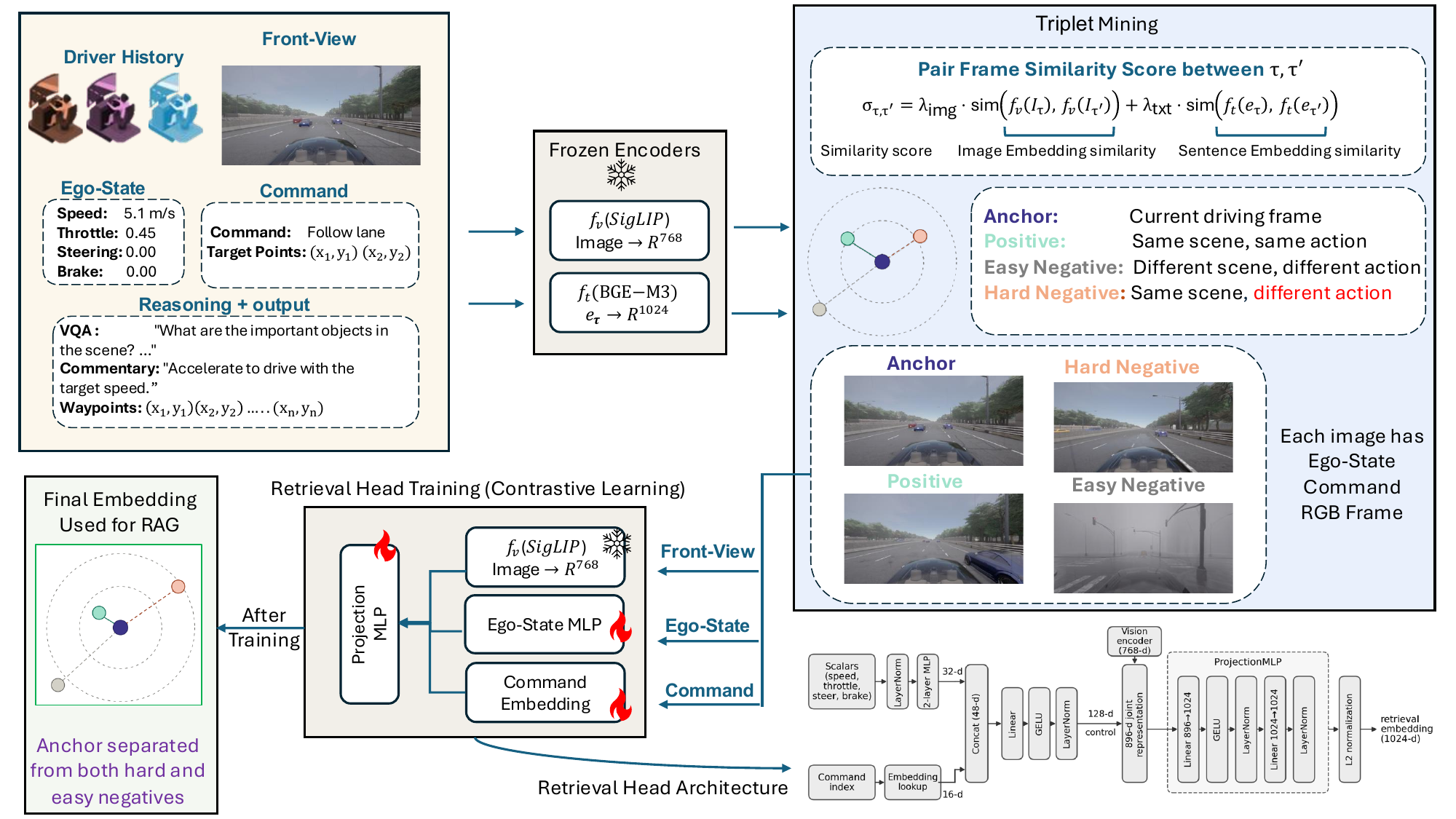}
 \caption{\textbf{Stage 1 triplet mining and Stage 2 retrieval head training.} Frozen SigLIP and BGE-M3 encoders score each pair of frames with the combined image and text similarity of Eq.~\ref{eq:combined_sim}. For each anchor, a positive matches it in scene, command, reasoning, and action. A hard negative looks similar but has a different command or speed. An easy negative differs in both scene and action, and it gets the lowest score. Bottom: the retrieval head is trained on these triplets (middle, Section~\ref{sec:framework:retrieval}). Its layers are shown on the right, and the left shows the final embedding, which separates the anchor from both negatives.}
  \label{fig:triplet}
\end{figure*}

\begin{itemize}
    \item \textbf{Positives} $\xi^{p,+}$: the top-$N^{+}$ percentile of $\mathcal{H}^p$ ranked by $\sigma^p$, which agree with the anchor in scene, context, and resulting action.
    \item \textbf{Easy negatives} $\xi^{p,-}_{\text{easy}}$: the bottom-$N^{-}$ percentile by $\sigma^p$, which differ from the anchor in both scene and action.
    \item \textbf{Hard negatives} $\xi^{p,-}_{\text{hard}}$: drawn from the top image-similarity percentile $\mathrm{sim}(f_v(\mathbf{I}^p_\tau), f_v(\mathbf{I}^p_{\tau'}))$ but restricted to frames whose navigational command or speed disagrees with the anchor's. They usually score just below the positives in $\sigma^p$ and well above the easy negatives, yet led to different waypoints. A retrieval head using visual features alone would rank them as positives, which makes them the most informative training signal.
\end{itemize}

\subsection{Stage 2: Retrieval Head Training}
\label{sec:framework:retrieval}

Stage~2 trains the retrieval head used at inference time. It uses only visual features and control signals, so embedding a frame needs no sentence or language-model pass. Database frames reuse the SigLIP features from Stage~1, and a query costs one frozen SigLIP pass plus two small MLPs. A single retrieval head is shared across all three styles, and only the index it queries changes.

\textbf{Retrieval embedding.}
Each frame $\tau$ in style history $\mathcal{H}^p$ is represented by concatenating the precomputed frozen SigLIP image embedding $f_v(\mathbf{I}^p_\tau) \in \mathbb{R}^{d_v}$ with a control embedding produced by a trained ControlEncoder $f_c$, and projecting the result through a trained MLP $f_{\text{ret}}$:
\begin{equation}
\small
\label{eq:retrieval_embedding}
\mathbf{s}^p_\tau \;=\; L_2\text{-norm}\!\left( f_{\text{ret}}\!\left[\, f_v(\mathbf{I}^p_\tau) \,\big\Vert\, f_c(\mathbf{u}^p_\tau,\, b^p_\tau,\, c^p_\tau) \,\right] \right) \;\in\; \mathbb{R}^{d_r},
\end{equation}
where
$\mathbf{u}^p_\tau = (v^p_\tau,\ \text{throttle}^p_\tau,\ \delta^p_\tau)^\top$ holds the speed, throttle, and steering, $b^p_\tau \in \{0, 1\}$ is the brake flag, and $c^p_\tau$ is the navigational command. The encoder $f_c$ embeds the four control values with a small MLP and the command with a learned embedding. The vision encoder $f_v$ stays frozen, while $f_c$ and $f_{\text{ret}}$ are trained. The image embedding has $d_v = 768$ dimensions and the control embedding has $128$. Their concatenation goes through a two-layer MLP and is $L_2$-normalized to $d_r = 1024$ dimensions. A full layer-by-layer description is in Supp.~Sec.~\suppref{app:retrieval_arch}.

\textbf{Weighted triplet loss.}
We train $f_c$ and $f_{\text{ret}}$ together on the Stage~1 triplets. The triplets come from all three style datasets and from a $5\%$ slice of SimLingo's training data~\cite{renz2025simlingo}, which is small but spans many different scenario types. That data was recorded by PDM-lite, a rule-based expert driver, and carries no style. The result is one retrieval head for both the no-style and the per-style settings. Let $\mathbf{s}^{p,a}$, $\mathbf{s}^{p,+}$, and $\mathbf{s}^{p,-}$ be the embeddings (Eq.~\ref{eq:retrieval_embedding}) of the anchor, positive, and negative frames of style $p$. The base triplet loss is
\begin{equation}
\small
\label{eq:base_triplet}
\mathcal{L}^p_{\text{tri}} = \max\!\left( \| \mathbf{s}^{p,a} - \mathbf{s}^{p,+} \|_2 - \| \mathbf{s}^{p,a} - \mathbf{s}^{p,-} \|_2 + \beta,\ 0 \right),
\end{equation}
where $\beta > 0$ is the margin. Hard negatives receive a higher weight $w_h > 1$ to penalize the model more for confusing frames that are visually similar but involve different driving context:
\begin{equation}
\small
\label{eq:weighted_triplet}
\mathcal{L}^p_{\text{ret}} \;=\; w \cdot \mathcal{L}^p_{\text{tri}}, \qquad
w = \begin{cases}
w_h & \text{if } \xi^{p,-} = \xi^{p,-}_{\text{hard}} \\
1   & \text{if } \xi^{p,-} = \xi^{p,-}_{\text{easy}}
\end{cases}.
\end{equation}
The total retrieval training loss averages Eq.~\ref{eq:weighted_triplet} over the
partitions of the pooled triplet set, namely the three styles and the
style-agnostic PDM-lite slice:
$\mathcal{L}_{\text{ret}} = \frac{1}{|\mathcal{P}'|} \sum_{p \in \mathcal{P}'} \mathbb{E}_\tau [\mathcal{L}^p_{\text{ret}}]$, where
$\mathcal{P}' = \mathcal{P} \cup \{\text{style-agnostic}\}$.

\textbf{Style databases and inference-time retrieval.}
Once trained, $f_c$ and $f_{\text{ret}}$ encode every retained frame of each style history offline. Each frame is stored together with its context point $\xi^p_\tau$ (defined in Section~\ref{sec:framework:context_point}) in a per-style FAISS~\cite{johnson2017faiss} index, producing three indices, one per style. At inference, the query embedding for the current observation at step $t$ (the current frame) is computed identically:
\begin{equation}
\small
\label{eq:query_embedding}
\mathbf{s}^q_t \;=\; L_2\text{-norm}\!\left( f_{\text{ret}}\!\left[\, f_v(\mathbf{I}_t) \,\big\Vert\, f_c(\mathbf{u}_t,\, b_t,\, c_t) \,\right] \right) \;\in\; \mathbb{R}^{d_r},
\end{equation}
and retrieval selects the set $\mathcal{C}^p_t$ of the top-$K$ most similar context points from the style-$p$ database, with $K = 2$ here:
\begin{equation}
\small
\label{eq:topk_retrieval}
\mathcal{C}^p_t \;=\; \Big\{\, \xi^p_\tau \;:\; \tau \in \operatorname*{arg\,top-\mathit{K}}_{\tau \in \{1,\dots,T^p\}} \; \mathrm{sim}\!\left( \mathbf{s}^q_t,\ \mathbf{s}^p_\tau \right) \Big\},
\end{equation}
where $T^p = |\mathcal{H}^p|$ is the number of context points stored in the style-$p$ database. Switching the agent's style at inference is therefore a single operation: change which of the three FAISS indices the query is issued against. No parameter is updated.

\subsection{Context-Point Tuple}
\label{sec:framework:context_point}

Each entry in the style-$p$ database is a context point $\xi^p_\tau = (\mathbf{I}^p_\tau,\, \mathbf{Q}^p_{\tau-2:\tau},\, c^p_\tau,\, \mathbf{g}^p_\tau,\, r^p_\tau,\, \mathbf{W}^{p,\star}_\tau)$ with six fields. It holds the front-view RGB frame $\mathbf{I}^p_\tau$ and a short control history $\mathbf{Q}^p_{\tau-2:\tau} \in \mathbb{R}^{2 \times 3}$. The control history stores the speed and throttle at the anchor frame and at the two frames before it. The context point also holds the navigational command $c^p_\tau$, the next two GPS target points $\mathbf{g}^p_\tau \in \mathbb{R}^{2 \times 2}$, and the commentary $r^p_\tau$. Finally, it stores the $N = 10$ waypoints $\mathbf{W}^{p,\star}_\tau \in \mathbb{R}^{N \times 2}$ driven under style $p$ after frame $\tau$. Supp.~Sec.~\suppref{app:context_point} discusses the different roles these fields play during mining, retrieval, and prompting.

\subsection{Stage 3: Prompt Structure and Supervised Fine-Tuning}
\label{sec:framework:sft}

\textbf{Prompt construction.}
Once retrieval returns $\mathcal{C}^p_t$, we build a prompt $\mathcal{X}_t$ that places the $K = 2$ retrieved demonstrations before the current observation. The backbone thus sees the demonstrations as in-context examples and predicts waypoints for the current scene by analogy:
\begin{equation}
\small
\label{eq:prompt}
\begin{aligned}
\mathcal{X}_t \;=\;\;
    & \underbrace{\,\xi^p_{(1)} \,\big\Vert\, \xi^p_{(2)}\,}_{\text{(1) retrieved demonstrations}}
      \,\Big\Vert\,
      \underbrace{\,[\, \mathbf{I}_t,\, v_t,\, c_t,\, \mathbf{g}_t \,]\,}_{\text{(2) current observation}} \\[8pt]
    & \Big\Vert\,
      \underbrace{\,\text{QUESTION}\,}_{\text{(3) ``What should the ego do next?''}}.
\end{aligned}
\end{equation}
Each retrieved context point $\xi^p_{(k)} \in \mathcal{C}^p_t$ is turned into a fixed sequence of tokens, in the same order as the in-context template of the backbone:
\begin{equation}
\small
\label{eq:serialize}
\begin{aligned}
\mathrm{serialize}\!\left( \xi^p_{(k)} \right) = \big(\,
    & \mathbf{I}^p_{(k)},\ \mathbf{Q}^p_{(k)},\ c^p_{(k)}, \\
    & \mathbf{g}^p_{(k)},\ r^p_{(k)},\ \mathbf{W}^{p,\star}_{(k)} \,\big).
\end{aligned}
\end{equation}
The prompt therefore contains the six fields of the context point (Section~\ref{sec:framework:context_point}), in the order of Eq.~\ref{eq:serialize}. The VQA annotation is not one of these fields, so it never enters the prompt. It is used only for mining in Stage~1, since the commentary already describes the reasoning behind each demonstrated action. The backbone's own InternViT encoder turns the image into language-model tokens, and all other fields are written as text. Two vision encoders thus run at every step. SigLIP embeds the current frame for retrieval, and InternViT encodes the images that enter the prompt. Decoding has two steps. The backbone first writes the commentary as its answer to the question. Learned query tokens are then appended, $20$ for the geometric waypoints and $10$ for the temporal waypoints. The Route MLP heads read the language model's hidden states at these positions, so both sets of waypoints come from the same token stream in one extra forward pass. A worked example of a full prompt is in Supp.~Sec.~\suppref{app:prompt_example}.

\textbf{Supervised fine-tuning (SFT).}
A pre-trained VLA backbone does not know how to use retrieved context points as demonstrations. Without adaptation, it treats in-context examples like any other tokens, and they barely change its output~\cite{yuan2024ragdriver}. This stage therefore teaches the backbone to \emph{read} the $K$ retrieved demonstrations as few-shot examples and to imitate the driving they show. The fine-tuning data has two sources. \emph{All} PersonaDrive style demonstrations, from the three styles, add behavioral variety. A $20\%$ subset of the PDM-lite data keeps the basic driving skill. We split this mixture $80/20$ into training and validation at the route level, and the Bench2Drive evaluation routes appear in neither split. In each training example, the prompt $\mathcal{X}_t$ is paired with the ground-truth geometric and temporal waypoints $\mathbf{W}^{p,\star,\text{pos}}_t$ and $\mathbf{W}^{p,\star,\text{vel}}_t$ of the current frame, and the retrieved context points come from the same style database as these targets. We update the language model through a LoRA adapter~\cite{hu2022lora} and also train the projector, driving adaptor, and waypoint encoder (Supp.~Sec.~\suppref{app:hyperparameters}), while the vision encoder and the retrieval head stay frozen.

No style label is given. The model must infer the intended style from the retrieved demonstrations and reflect it in its output. We keep SimLingo's training objective~\cite{renz2025simlingo} unchanged. It supervises both outputs, with a smooth-$L_1$ loss on the geometric and temporal waypoints and a cross-entropy loss on the language tokens. All terms have equal weight, and the loss is averaged over all training examples. The full expression is in Supp.~Sec.~\suppref{app:sft_loss}.

\section{Experiments}
\label{sec:experiments}

Section~\ref{sec:experiments:setup} describes the data, benchmark, and
training configuration shared by all runs. We then evaluate closed-loop
driving without style conditioning (Section~\ref{sec:exp:b2d_base}),
ablate retrieval depth (Section~\ref{sec:exp:ablation}), and evaluate
style-conditioned driving against style-conditioning baselines
(Section~\ref{sec:exp:b2d_style}).

\subsection{Experimental Setup}
\label{sec:experiments:setup}

\paragraph{Data collection.}
We collected style data from $15$ participants on a driving simulator connected to CARLA. They drove $55$ CARLA Leaderboard~2.0 routes covering $21$ scenario types. The simulator runs at $20$\,Hz, and we save every fifth step, so the $29{,}777$ recorded frames are $0.25$\,s apart. Mining then keeps every fifth recorded frame (Section~\ref{sec:framework:triplets}). Figure~\ref{fig:data_collection} summarizes the protocol and the recorded signals. Each participant drives their own set of routes three times, in random order, so that style is not confused with familiarity with the route. Each time, they follow one of three instructions. \emph{Conservative} asks them to keep a low target speed and to yield whenever possible. \emph{Neutral} asks them to follow traffic laws and drive safely at a reasonable speed. \emph{Aggressive} asks them to drive as fast as possible without crashing, and to tailgate and cut off merging vehicles. On the same routes, the mean speed rises from $9.8$\,mph under the conservative instruction to $13.2$\,mph under the neutral one and $17.5$\,mph under the aggressive one, so the aggressive drives are $79\%$ faster than the conservative ones. The instructions therefore produced clearly different behavior before any model is involved. Supp.~Sec.~\suppref{app:data_collection} describes the rig, the exact instructions given to the participants, the speed statistics, and the annotations.

\textbf{Benchmark and metrics.}
We evaluate on the Bench2Drive closed-loop benchmark~\cite{jia2024bench2drive}, which has $220$ routes in Town01--Town15 with $44$ interactive scenarios from CARLA Leaderboard~2.0. We report its official metrics. The \emph{Driving Score} (DS) is the main one. It combines how much of each route the agent completes with a penalty for every infraction, such as a collision or a red-light violation. The \emph{Success Rate} (SR) is the percentage of routes completed without any infraction. \emph{Efficiency} compares the agent's speed with the speed of the surrounding traffic, and \emph{Comfort} measures how smooth the driving is. We also report the average speed and longitudinal acceleration. For open-loop accuracy, we use the average displacement error (ADE), which is the mean distance in meters between the predicted and the recorded waypoints. The final displacement error (FDE) is the same distance at the last waypoint. Our collection routes use scenario types that also appear in Bench2Drive, but none of them is one of its $220$ test routes. Retrieval therefore never returns a frame from a route on which the agent is evaluated.

\textbf{Model and training configuration.}
The VLA backbone is fine-tuned with the in-context SFT of Section~\ref{sec:framework:sft} (Stage~3), using a LoRA adapter on the language model. The vision encoder and the retrieval head stay frozen. Every PersonaDrive result below comes from this single Stage~3 checkpoint. It was trained on the $20\%$ PDM-lite slice and on our style demonstrations, with the $80/20$ route-level split of Section~\ref{sec:framework:sft}. On top of the backbone, we add the retrieval head (Supp.~Sec.~\suppref{app:retrieval_arch}) and the prompt format of Section~\ref{sec:framework:sft}, and only the searched index changes between experiments. Each query retrieves $K = 2$ demonstrations from the selected index. Section~\ref{sec:exp:ablation} varies $K$ and explains this choice through the trade-off between accuracy and speed. The adapter rank, learning rates, and all other hyperparameters of the three training stages are in Supp.~Sec.~\suppref{app:hyperparameters}.

\subsection{Experiment 1: Closed-Loop Driving on Bench2Drive (No Style)}
\label{sec:exp:b2d_base}

\textbf{Goal and training data.}
This experiment asks whether retrieval keeps, harms, or improves closed-loop driving compared with SimLingo. No style is used in this experiment. Queries go to a style-agnostic FAISS index built from a $5\%$ route-level subset of SimLingo's PDM-lite training data~\cite{renz2025simlingo}, embedded by the same retrieval head (Section~\ref{sec:framework:retrieval}). This subset, like the $20\%$ subset used for fine-tuning, covers the CARLA Leaderboard scenario types~\cite{carla_lb1_scenarios} that Bench2Drive tests (Supp.~Sec.~\suppref{app:training_data}). PersonaDrive therefore differs from SimLingo in only two ways. It retrieves demonstrations, and it receives extra fine-tuning in Stage~3. Any gain over SimLingo must therefore come from one of these two. To find out which, we also fine-tune SimLingo on the same data without retrieval and report it in the results below.

\textbf{Results.}
Table~\ref{tab:b2d_base} compares PersonaDrive with the methods reported in the SimLingo paper~\cite{renz2025simlingo} and with HiP-AD~\cite{tang2025hipad}. Without any style conditioning, PersonaDrive has the highest Driving Score and Success Rate of all methods in Table~\ref{tab:b2d_base}. Its Driving Score is $4.6\%$ higher than SimLingo's and $2.5\%$ higher than that of HiP-AD, the strongest prior method in the table. Its Success Rate is $7.5\%$ and $4.6\%$ higher, so it finishes more routes without any infraction, such as an accident. For example, Supp.~Fig.~\suppref{fig:agent_behavior} shows PersonaDrive stopping at a red light that SimLingo drives through. We then examine whether this gain stems from retrieval or merely from additional fine-tuning. To do so, we fine-tune SimLingo on the same routes with the same settings, but without retrieved demonstrations (\emph{SimLingo + SFT, no retrieval} in Table~\ref{tab:b2d_base}). This model scores close to SimLingo. Fine-tuning alone therefore does not explain the gain, which comes from the retrieved demonstrations. Each retrieved demonstration shows how a comparable scene, with a similar road layout, command, and traffic, was actually driven. Where infractions tend to happen, the backbone thus has a concrete example to follow rather than only the current frame. This gain in safety comes with somewhat less smooth driving. Efficiency is $1.6\%$ lower than SimLingo's, and Comfort is $17\%$ lower.
\begin{table}[tb]
\centering
\caption{\textbf{Closed-loop results on Bench2Drive without style conditioning.} PersonaDrive searches the style-agnostic index of Section~\ref{sec:exp:b2d_base}, and its training data is the same as in Section~\ref{sec:experiments:setup}. Prior-work numbers are from~\cite{renz2025simlingo} and~\cite{tang2025hipad}. Bold marks the best value in each column. \emph{SimLingo + SFT, no retrieval} is SimLingo fine-tuned on the same routes and settings as PersonaDrive but without retrieval (one evaluation run). It is not ranked for bold.}
\label{tab:b2d_base}
\small
\setlength{\tabcolsep}{3pt}
\begin{adjustbox}{max width=\linewidth}
\begin{tabular}{lcccc}
\toprule
\textbf{Method} & DS\,$\uparrow$ & SR\,(\%)\,$\uparrow$ & Eff.\,$\uparrow$ & Comf.\,$\uparrow$ \\
\midrule
\multicolumn{5}{l}{\textit{Think2Drive~\cite{li2024think2drive} expert (with distillation)}} \\
TCP~\cite{wu2022tcp}                & 40.70 & 15.00 & 54.26  & 47.80 \\
TCP-ctrl                            & 30.47 &  7.27 & 55.97  & \textbf{51.51} \\
TCP-traj                            & 59.90 & 30.00 & 76.54  & 18.08 \\
ThinkTwice~\cite{jia2023thinktwice} & 62.44 & 31.23 & 69.33  & 16.22 \\
DriveAdapter~\cite{jia2023driveadapter} & 64.22 & 33.08 & 70.22  & 16.01 \\
\midrule
\multicolumn{5}{l}{\textit{Think2Drive expert (without distillation)}} \\
AD-MLP~\cite{zhai2023admlp}         & 18.05 &  0.00 & 48.45  & 22.63 \\
UniAD-Tiny~\cite{hu2023uniad}       & 40.73 & 13.18 & 123.92 & 47.04 \\
UniAD-Base~\cite{hu2023uniad}       & 45.81 & 16.36 & 129.21 & 43.58 \\
VAD~\cite{jiang2023vad}             & 42.35 & 15.00 & 157.94 & 46.01 \\
HiP-AD~\cite{tang2025hipad}         & 86.77 & 69.09 & 203.12 & 19.36 \\
\midrule
\multicolumn{5}{l}{\textit{PDM-lite expert (open-source)}} \\
TCP-traj w/o distill                & 49.30 & 20.45 & 78.78   & 22.96 \\
SimLingo-BASE~\cite{renz2025simlingo}  & 85.94 & 66.82 & 244.18 & 25.49 \\
SimLingo~\cite{renz2025simlingo}    & 85.07\,$\pm$\,0.95 & 67.27\,$\pm$\,2.11 & \textbf{259.23\,$\pm$\,5.59} & 33.67\,$\pm$\,5.72 \\
\midrule
SimLingo + SFT, no retrieval & 85.08 & 66.36 & 259.34 & 30.83 \\
\rowcolor{oursrow}\textbf{PersonaDrive (no style)}    & \textbf{88.95} & \textbf{72.29} & 255.15 & 28.09 \\
\bottomrule
\end{tabular}
\end{adjustbox}
\end{table}

\subsection{Ablation: Retrieval Depth}
\label{sec:exp:ablation}

\begin{table}[tb]
\centering
\caption{\textbf{Effect of the number of retrieved demonstrations $K$ (no style).} $K = 0$ is the SimLingo backbone without retrieval. $K \geq 1$ uses the PersonaDrive backbone, so $K = 2$ matches the PersonaDrive row of Table~\ref{tab:b2d_base}. ADE and FDE are open-loop errors, and DS, SR, Eff., and Comf.\ are Bench2Drive closed-loop metrics. Tokens is the prompt length and Total the end-to-end time per step. The full breakdown is in Supp.~Sec.~\suppref{app:k-depth}.}
\label{tab:k-depth-main}
\small
\setlength{\tabcolsep}{3pt}
\begin{adjustbox}{max width=\linewidth}
\begin{tabular}{lcc>{\columncolor{oursrow}}ccc}
\toprule
\textbf{Metric} & $K{=}0$ & $K{=}1$ & $K{=}2$ & $K{=}3$ & $K{=}4$ \\
\midrule
ADE (m)\,$\downarrow$ & 0.820 & 0.880 & 0.795 & 0.793 & 0.791 \\
FDE (m)\,$\downarrow$ & 1.787 & 1.961 & 1.699 & 1.650 & 1.648 \\
\midrule
DS\,$\uparrow$        & 85.07 & 81.55 & 88.95 & 89.16 & 89.30 \\
SR\,(\%)\,$\uparrow$  & 67.27 & 62.17 & 72.29 & 72.35 & 73.10 \\
Eff.\,$\uparrow$      & 259.23 & 259.45 & 255.15 & 261.82 & 261.01 \\
Comf.\,$\uparrow$     & 33.67 & 25.77 & 28.09 & 30.24 & 33.80 \\
\midrule
Tokens\,$\downarrow$     & 547 & 1292 & 2030 & 2769 & 3507 \\
Total (ms)\,$\downarrow$ & 480.3 & 509.6 & 557.7 & 599.4 & 649.7 \\
\bottomrule
\end{tabular}
\end{adjustbox}
\end{table}

\paragraph{Choice of $K$.}
This ablation asks how many retrieved demonstrations the backbone needs and what they cost. We vary $K$ from $0$ to $4$. $K = 0$ is the original SimLingo backbone without retrieval, so this column is a reference point rather than part of the controlled sweep. For $K \geq 1$, the checkpoint and the index stay the same, so only the retrieved context changes. ADE and FDE are measured on held-out PDM-lite routes that are not in the $5\%$ and $20\%$ training slices. Table~\ref{tab:k-depth-main} shows that open-loop and closed-loop accuracy improve together from $K = 1$ to $K = 4$. The result at $K = 1$ shows how much the fine-tuned backbone relies on retrieval. With a single demonstration, its Driving Score is $4.1\%$ lower than at $K = 0$, and its ADE and FDE are $7\%$ and $10\%$ worse. With two demonstrations, the same fine-tuned backbone scores $4.6\%$ above $K = 0$. Beyond $K = 2$, each added demonstration makes the prompt about $740$ tokens longer. At $K = 2$, a step takes $16\%$ longer than SimLingo's, and only a quarter of this extra time is the FAISS lookup. Since closed-loop driving pays this cost at every step, we use $K = 2$. This choice trades some smoothness for speed, because Efficiency and Comfort are higher at $K = 4$, while the Driving Score barely changes.

\begin{table}[tb]
\centering
\caption{\textbf{Bench2Drive closed-loop results under three style instructions.} Baseline numbers are from~\cite{wang2026drivemyway}, and PersonaDrive follows the same protocol. SimLingo is prompted with the style name here, so its scores differ from the unprompted result in Table~\ref{tab:b2d_base}. Bold marks the best value per style. Speed and acceleration have no preferred direction, so they are not bolded.}
\label{tab:bench2drive}
\begin{adjustbox}{max width=\linewidth}
\begin{tabular}{l|l|cccccc}
\toprule
\textbf{Method} & Style & DS\,$\uparrow$ & SR\,(\%)\,$\uparrow$ & Eff.\,$\uparrow$ & Comfort\,$\uparrow$ & Speed (m/s) & Accel. (m/s$^2$) \\
\midrule
\multirow{3}{*}{SimLingo~\cite{renz2025simlingo}}
    & Conservative & 78.18 & 65.56 & 238.77 & 26.99  & 7.21 & 5.29 \\
    & Neutral      & 78.15 & 65.85 & 241.44 & 24.67  & 7.37 & 5.22 \\
    & Aggressive   & 78.56 & 65.83 & 247.60 & 18.61  & 7.66 & 5.39 \\
\midrule
\multirow{3}{*}{StyleDrive~\cite{hao2026styledrive}}
    & Conservative & 77.02 & 61.96 & 242.18 & 23.67  & 6.82 & 5.39 \\
    & Neutral      & 76.26 & 62.13 & 249.07 & 21.35  & 6.98 & 5.43 \\
    & Aggressive   & 75.68 & 60.89 & 256.71 & 16.79  & 7.23 & 5.59 \\
\midrule
\multirow{3}{*}{DMW~\cite{wang2026drivemyway}}
    & Conservative & 82.72 & 71.56 & 237.06 & \textbf{34.62}  & 6.18 & 5.26 \\
    & Neutral      & 82.03 & \textbf{70.95} & 244.98 & 28.67  & 6.34 & 5.43 \\
    & Aggressive   & 79.50 & 67.36 & \textbf{281.56} & 21.62  & 7.72 & 6.01 \\
\midrule
\rowcolor{oursrow}
    & Conservative & \textbf{89.62} & \textbf{73.57} & \textbf{258.44} & 33.05  & 5.05 & 3.17 \\
\rowcolor{oursrow}
    & Neutral      & \textbf{87.55} & 69.50 & \textbf{264.83} & \textbf{31.20}  & 5.52 & 3.51 \\
\rowcolor{oursrow}
\multirow{-3}{*}{\textbf{PersonaDrive (Ours)}}
    & Aggressive   & \textbf{88.69} & \textbf{72.56} & 269.20 & \textbf{27.85}  & 6.05 & 3.96 \\
\bottomrule
\end{tabular}
\end{adjustbox}
\end{table}

\subsection{Experiment 2: Style-Conditioned Closed-Loop Evaluation}
\label{sec:exp:b2d_style}

\paragraph{Goal and baselines.}
We now run PersonaDrive under each of the three style instructions. We ask whether searching the per-style databases produces driving that differs between styles in the intended way. The backbone is the same Stage~3 checkpoint as in Experiment~1, and only the searched index changes between styles. We follow the evaluation protocol of Drive My Way (DMW)~\cite{wang2026drivemyway}. We compare with three style-conditioning baselines. SimLingo~\cite{renz2025simlingo} is prompted with the style name in natural language. A StyleDrive-like baseline, implemented in~\cite{wang2026drivemyway} after StyleDrive~\cite{hao2026styledrive}, maps each style instruction to a style condition. DMW adapts SimLingo to each style with GRPO fine-tuning, using reward weights that an LLM infers for each scenario.

\textbf{Results.}
Table~\ref{tab:bench2drive} shows that PersonaDrive has the highest Driving Score in every style. A higher Driving Score means that the agent finishes more of each route with fewer infractions, such as collisions or red-light violations. Its margin over DMW, the strongest baseline, ranges from $6.7\%$ for the neutral style to $11.6\%$ for the aggressive style. Its three styles also stay within $2.4\%$ of each other, so controlling the style does not hurt driving quality. Its Success Rate is close to DMW's. It is higher for the conservative and aggressive styles and $1.45$ points lower for the neutral style, a gap smaller than SimLingo's $\pm 2.11$-point spread between runs (Table~\ref{tab:b2d_base}).
The style metrics also follow the instructions. From the conservative to the aggressive style, average speed rises by $19.8\%$ and acceleration by $24.9\%$, and efficiency rises as well. Comfort drops by $15.7\%$, as expected for faster and less smooth driving. Prompting SimLingo with the style name changes its behavior much less. Its speed rises by only $6.2\%$ from conservative to aggressive, its acceleration shows no clear order, and its Driving Score changes by less than $1\%$. The videos on the project page (Supp.~Sec.~\suppref{app:data_collection}) show the same difference in single decisions. On one route, the aggressive agent overtakes a cyclist right away, while the conservative agent waits behind it.

\section{Conclusion and Future Work}
We presented PersonaDrive, a VLA driving agent that takes its driving style from retrieved demonstrations of human drivers instead of fixed parameters or proxy rewards. Changing the style only requires searching a different index, so no retraining is needed. On Bench2Drive without style conditioning, PersonaDrive improves the Driving Score by $4.6\%$ over SimLingo and by $2.5\%$ over HiP-AD. With style conditioning, it leads in every style, and its speed and acceleration follow the instruction. Our study has a limitation. We evaluate each agent alone, as the ego vehicle. We do not yet study how several agents with different styles interact in one scene, including agents controlled by real human drivers. Future work will therefore run several style-conditioned agents in one environment and let a human drive among them.

\section*{Acknowledgements}

This work is supported by the U.S. National Science Foundation (NSF) under grant number 2339266 and is partially supported by the UCI ProperAI Institute, an Engineering+Society Institute funded as part of a generous gift from Susan and Henry Samueli.

{
    \small
    \bibliographystyle{ieeenat_fullname}
    \bibliography{references}
}

\clearpage
\appendix
\setcounter{figure}{0}\setcounter{table}{0}\setcounter{equation}{0}
\renewcommand{\thefigure}{S\arabic{figure}}
\renewcommand{\thetable}{S\arabic{table}}
\renewcommand{\theequation}{S\arabic{equation}}
\appendix
\iftoggle{inlineappendix}{\newpage\section*{Supplementary Material}}{}

This supplementary material complements the main paper, starting with an extended discussion of the related work in Sec.~\ref{app:related_work}, which also compares recent open-loop and closed-loop driving methods in a table. Sec.~\ref{app:model_description} describes the whole system, Sec.~\ref{app:retrieval_arch} gives the layer-by-layer architecture of the retrieval head and shows examples of mined triplets, and Sec.~\ref{app:frame_selection} explains why only every fifth frame is kept for triplet mining. Sec.~\ref{app:data_collection} covers the data collection, namely the driving rig, the collection interface, the style instructions, the choice of routes, which are disjoint from the Bench2Drive evaluation routes, the speed statistics that show the participants followed the instructions, participant consent, recording and annotation, and two qualitative examples of the resulting driving behavior. Sec.~\ref{app:training_data} explains how we sampled the PDM-lite routes and which scenarios they cover. Sec.~\ref{app:hyperparameters} lists the hyperparameters, the compute, and how the hyperparameters were chosen. Sec.~\ref{app:context_point} formalizes the context point, Sec.~\ref{app:k-depth} reports the full retrieval-depth results with the time per step, Sec.~\ref{app:sft_loss} states the fine-tuning loss, Sec.~\ref{app:prompt_example} shows a worked example of the prompt, and Sec.~\ref{app:notation} collects the notation of the main paper and this supplement. Equations, figures, and tables without the S prefix refer to the main paper.

\section{Extended Related Work}
\label{app:related_work}

This section discusses in more detail the two lines of work summarized in the main paper. The first, building capable driving agents in simulation, covers traffic-agent control and end-to-end VLA models for driving. The second, style conditioning and retrieval grounding, covers style-conditioned and personalized driving and retrieval-augmented driving.

\paragraph{Traffic agent control in simulation.}
Realistic driving simulation needs non-ego traffic agents whose behavior is as varied as that of real human drivers. Rule-based traffic managers, such as CARLA's TrafficManager~\cite{dosovitskiy2017carla} and SUMO~\cite{lopez2018sumo}, produce agents that avoid collisions and obey the law. However, all their agents share a single behavioral mode. Learned approaches add variety across agents through reinforcement learning~\cite{shiroshita2020behaviorally}, autoregressive prediction~\cite{zhou2024behaviorgpt}, and LLM-guided hierarchical behavior models~\cite{li2026diverse}. However, these methods define style through engineered parameters or statistical proxies rather than grounding it in how humans actually behave under specific style instructions. PersonaDrive instead builds each agent as a full reasoning ego agent that, at inference time, follows retrieved demonstrations of real humans driving in the target style. Its behavioral variety therefore comes from reasoning grounded in how individual human drivers handled similar situations, rather than from changed parameters. In this work, we evaluate each agent on its own, as the ego vehicle. Such agents could later help to train ego policies in simulators like CARLA before they are transferred to the real world.

\paragraph{End-to-end VLA for autonomous driving.}
Recent VLA models have established strong foundations for end-to-end driving, because they combine perception, language reasoning, and action prediction. DriveLM~\cite{sima2024drivelm} links perception to planning through graph-structured visual question answering (VQA), and SimLingo~\cite{renz2025simlingo} reaches state-of-the-art closed-loop performance with action dreaming and chain-of-thought commentary. FeD~\cite{zhang2024feedback} refines its waypoint predictions with language feedback, AutoVLA~\cite{zhou2025autovla} combines autoregressive action generation with GRPO fine-tuning for adaptive reasoning, and Alpamayo-R1~\cite{wang2025alpamayo} pairs language reasoning with a diffusion-based trajectory decoder and reinforcement-learning post-training. These models show that VLA backbones can be capable ego agents in simulation, but each produces a single learned behavior rather than a controllable range of driving styles. PersonaDrive builds directly on SimLingo's backbone and adds style control through retrieval and in-context learning.

\paragraph{Style-conditioned and personalized driving.}
StyleDrive~\cite{hao2026styledrive} contributes a benchmark for style-aware driving whose baselines receive the style as a categorical token. This token fixes, at training time, the behavior that a style produces, and its labels are inferred after the fact from everyday driving rather than from drivers instructed to adopt a style. MAVERIC~\cite{schrum2024maveric} fits one style embedding per person from driving data and questionnaire answers, so it targets one person at a time rather than a population. Drive My Way~\cite{wang2026drivemyway} aligns SimLingo to individual preferences. It infers a reward weighting per scenario with an LLM, has domain experts review it, and then runs GRPO. It also collects a dataset of $30$ drivers in CARLA and learns an embedding of each driver's long-term preferences from their driving and a questionnaire. However, these drivers drive naturally rather than under an explicit style instruction, so the dataset does not show how one person changes their driving when asked to adopt a different style. Driving Style Alignment~\cite{driving_style_alignment} replaces this reward model with repeated feedback from an LLM coach. However, its supervision is still an LLM's description of the style, not a record of it. Other work generates synthetic style data, either by editing expert trajectories to drive faster or slower (D$^3$-MoE~\cite{feng2026d3moe}) or by having a rule-based expert drive at chosen target speeds (Bench2Drive-Speed~\cite{shao2026bench2drivespeed}). All of these methods therefore learn style from a proxy, and none records how people drive when they are instructed to adopt a style. PersonaDrive instead treats a style as a class of behavior, grounded in what drivers actually do when told to drive aggressively, neutrally, or conservatively. This motivates our dataset, in which each driver drives each of their routes three times, with an explicit style instruction for each drive.

\paragraph{Retrieval-augmented driving.}
RAG-Driver~\cite{yuan2024ragdriver} and RealDrive~\cite{ding2025realdrive} ground driving decisions in retrieved past experiences and establish retrieval-augmented in-context learning as a workable approach for autonomous driving. However, both search a single shared database, so they cannot retrieve examples by style. RAG-Driver pairs a 7B video language model with a retrieval engine whose joint embedding of video and control signals is trained with a triplet loss, where positives are chosen by how similar the expert explanations are. It retrieves two examples from one database (BDD-X), writes its explanations and control signals as text, and is evaluated open-loop on recorded data. PersonaDrive differs in both architecture and setting. It keeps a separate database of human demonstrations for each style and mines its training triplets with a combined image and text similarity, where the text describes each frame through its VQA annotation, commentary, and driven waypoints. It also builds hard negatives from frames with a different command or speed. Finally, it feeds the retrieved demonstrations to a VLA whose waypoint heads drive the vehicle in closed-loop simulation.

\begin{table*}[t]
\centering
\caption{\textbf{Comparison of recent approaches} for open-loop and closed-loop autonomous driving, by input, alignment method, and evaluation protocol.}
\label{tab:related_work_comparison}
\small
\setlength{\tabcolsep}{4pt}
\begin{adjustbox}{max width=\textwidth}
\begin{tabular}{>{\raggedright\arraybackslash}p{2.0cm}%
                >{\raggedright\arraybackslash}p{3.3cm}%
                >{\raggedright\arraybackslash}p{2.0cm}%
                >{\raggedright\arraybackslash}p{3.5cm}%
                >{\raggedright\arraybackslash}p{2.0cm}%
                >{\raggedright\arraybackslash}p{2.9cm}}
\toprule
\textbf{Method} & \textbf{Core Approach} & \textbf{Inputs} & \textbf{Alignment Mechanism} & \textbf{Evaluation} & \textbf{Key Findings} \\
\midrule
DriveLM~\cite{sima2024drivelm}
    & Graph-structured VQA linking perception to planning
    & Multi-cam, Ego, Instr
    & Graph Q\&A with custom action tokens
    & Zero-shot E2E vs SOTA
    & Competitive E2E driving and gains in VQA \\
\addlinespace
SimLingo~\cite{renz2025simlingo}
    & Unified driving, language, and action alignment
    & Front cam, Ego, Instr
    & Action Dreaming with CoT commentary
    & CARLA LB 2.0, Bench2Drive
    & Strong closed-loop driving, and our backbone \\
\addlinespace
FeD~\cite{zhang2024feedback}
    & Language feedback to correct driving predictions
    & Front cam, Ego, Instr
    & Feedback-refined masked-token waypoint head
    & nuScenes, CARLA
    & High driving performance with distillation in the VLA \\
\addlinespace
Driving Style Alignment~\cite{driving_style_alignment}
    & Align LLM driver to human driving styles
    & Scene desc., Ego, Instr
    & Demonstration + coach feedback
    & CARLA with user studies
    & Distinct styles resembling humans \\
\addlinespace
AutoVLA~\cite{zhou2025autovla}
    & Autoregressive VLA with adaptive fast/slow reasoning
    & Multi-cam, Ego, Instr
    & Action codebook + GRPO fine-tuning
    & nuPlan, nuScenes, Waymo
    & Strong results, and fine-tuning reduces extra reasoning \\
\addlinespace
RAG-Driver~\cite{yuan2024ragdriver}
    & Retrieval-augmented in-context learning
    & Video, Instr
    & Retrieve similar driving scenes
    & BLEU, METEOR
    & Better explanations \& control predictions \\
\addlinespace
StyleDrive~\cite{hao2026styledrive}
    & Style-aware benchmarking of E2E driving with style-conditioned baselines
    & Cam, LiDAR, Ego (arch.-dependent)
    & One-hot categorical style tokens fused with trajectory query
    & StyleDrive benchmark
    & Style conditioning improves alignment \\
\addlinespace
Drive My Way~\cite{wang2026drivemyway}
    & Preference alignment via per-scenario LLM-inferred reward weights
    & Front cam, Ego, Profile
    & GPT-5-inferred reward weights with expert review, then GRPO fine-tuning
    & Bench2Drive
    & Strong style adaptation and OOD generalization via a frozen profile encoder \\
\addlinespace
\rowcolor{oursrow}\textbf{PersonaDrive (Ours)}
    & \textbf{Retrieval-guided style-conditioned control at run-time}
    & \textbf{Front cam, Ego, Instr}
    & \textbf{Per-style FAISS indices of human demonstrations, swapped at inference}
    & \textbf{Bench2Drive}
    & \textbf{Diverse human-style behaviors without backbone retraining} \\
\bottomrule
\end{tabular}
\end{adjustbox}
\end{table*}

\section{Model Description}
\label{app:model_description}

This section gives an end-to-end description of the PersonaDrive system, tying together the components introduced piecemeal in the main paper. PersonaDrive consists of three modules. The first is a \textbf{retrieval head} that turns a query frame into an embedding that reflects driving behavior. The second is a set of \textbf{per-style FAISS indices} filled with human demonstrations. The third is a \textbf{VLA backbone} (SimLingo, built on InternVL2-1B). It reads a prompt with the retrieved demonstrations and predicts waypoints, which PID controllers turn into control commands. All three styles share the same backbone weights and the same retrieval head. The only thing that changes between styles at inference is which per-style FAISS index is searched.

\paragraph{Retrieval head.}
The retrieval head maps the image and control signals of a context point to an $L_2$-normalized embedding $\mathbf{s}^p_\tau \in \mathbb{R}^{1024}$. The image features come from a frozen SigLIP vision encoder~\cite{zhai2023siglip} (applied to the front camera image of the ego vehicle, $f_v: \mathbb{R}^{H \times W \times 3} \to \mathbb{R}^{768}$). The control features come from a learned ControlEncoder $f_c$. It fuses a $16$-dimensional command embedding with a $32$-dimensional embedding of the speed, throttle, steering, and brake, computed by a LayerNorm and a two-layer MLP. The resulting $128$-dimensional control vector is concatenated with the image features. The ProjectionMLP $f_{\text{ret}}$, a two-layer MLP with GELU and LayerNorm, maps the result to $\mathbb{R}^{1024}$, and the output is then $L_2$-normalized. Only the ControlEncoder, including its fusion layer, and the ProjectionMLP are trained, while the SigLIP encoder stays frozen. Sec.~\ref{app:retrieval_arch} gives all input and output sizes.

\paragraph{Per-style FAISS indices.}
For each style $p \in \mathcal{P} = \{\text{conservative}, \text{neutral}, \text{aggressive}\}$, we keep a separate FAISS \texttt{IndexFlatIP} index. It holds the embeddings $\{\mathbf{s}^p_\tau\}_\tau$ of every kept context point $\xi^p_\tau$ recorded under that style. Because the embeddings have unit length, the inner-product search equals a cosine-similarity search. At inference, the style is chosen by choosing which index to search. The retrieval head itself does not depend on the style, but the pool of neighbors does. Each context point stores both its embedding and its six-field tuple from Eq.~\ref{eq:context_point}, namely the RGB frame, control history, command, target points, commentary, and driven waypoints. The VQA label used for mining is stored alongside it. The retrieved entries can thus go straight into the prompt without any further lookup.

\paragraph{VLA backbone.}
We use SimLingo's VLA backbone, built on InternVL2-1B, and keep its architecture unchanged, including the dual waypoint head and the PID controllers. The head predicts $20$ geometric waypoints $\mathbf{W}^{\text{pos}}_t \in \mathbb{R}^{20 \times 2}$, placed about $1$\,m apart along the path, and $10$ temporal waypoints $\mathbf{W}^{\text{vel}}_t \in \mathbb{R}^{10 \times 2}$, placed at $4$\,Hz over the next $2.5$\,s, so their spacing sets the speed. PersonaDrive places $K$ retrieved demonstrations $\{\xi^p_{(1)}, \dots, \xi^p_{(K)}\}$, serialized with Eq.~\ref{eq:serialize}, before the current observation (image, speed, command, and GPS targets) and the question of Eq.~\ref{eq:prompt}. The backbone is fine-tuned in Stage~3 with the waypoint and language loss of Eq.~\ref{eq:sft_loss}. No style label is given, so the backbone learns to infer the intended style from the retrieved demonstrations and to reflect it in the predicted ego waypoints.

\paragraph{Inference-time procedure.}
At each simulator step $t$, given the live observation $o_t$ and
a target style $p$, PersonaDrive executes:
\begin{enumerate}[leftmargin=1.2em,topsep=2pt,itemsep=1pt]
    \item \textbf{Query encoding.} Compute the query embedding
          $\mathbf{s}^q_t = L_2\text{-norm}(f_{\text{ret}}[\, f_v(\mathbf{I}_t) \,\Vert\, f_c(\mathbf{u}_t, b_t, c_t) \,])$
          via Eq.~\ref{eq:query_embedding}.
    \item \textbf{Retrieval.} Query the style-$p$ FAISS index for
          the top-$K$ nearest neighbors
          $\mathcal{C}^p_t = \{\xi^p_{(k)}\}_{k=1}^{K}$
          ($K = 2$ in our experiments) via
          Eq.~\ref{eq:topk_retrieval}.
    \item \textbf{Prompt serialization.} Construct the in-context
          prompt $\mathcal{X}_t$ by serializing the $K$ retrieved tuples
          followed by the live query observation, per
          Eq.~\ref{eq:prompt} and Eq.~\ref{eq:serialize}. A
          worked example of the serialized prompt is given in
          Sec.~\ref{app:prompt_example}.
    \item \textbf{Waypoint prediction.} Forward the prompt through
          the fine-tuned backbone to obtain
          $\hat{\mathbf{W}}_t = (\mathbf{W}^{\text{pos}}_t,
          \mathbf{W}^{\text{vel}}_t)$.
    \item \textbf{Control.} Decode the geometric waypoints through the lateral PID and the temporal waypoints through the longitudinal PID, inherited from SimLingo, to produce
          throttle, brake, and steering commands.
\end{enumerate}
Switching the style at run time therefore only means sending the search in step~2 to a different FAISS index. The backbone weights and the retrieval head are shared by all styles.

\paragraph{Three-stage training pipeline.}
The system is trained in three sequential stages, each with its
own loss and data slice:
\begin{itemize}[leftmargin=1.2em,topsep=2pt,itemsep=1pt]
    \item \textbf{Stage~1: Triplet mining.} We mine triplets (anchor, positive, negative) from the three per-style human demonstration databases and from a $5\%$ style-agnostic slice of SimLingo's PDM-lite training set. Mining uses the combined image and text similarity $\sigma^p_{\tau,\tau'}$ (Eq.~\ref{eq:combined_sim}). Both encoders are frozen. BGE-M3~\cite{chen2024bgem3} is used only for mining, while SigLIP is reused by the retrieval head at inference.
    \item \textbf{Stage~2: Retrieval head training.} We train the ProjectionMLP $f_{\text{ret}}$ and the ControlEncoder $f_c$ together on the mined triplets with the weighted triplet margin loss $\mathcal{L}_{\text{ret}}$ (Eq.~\ref{eq:weighted_triplet}). SigLIP stays frozen.
\item \textbf{Stage~3: In-context SFT of the backbone.} We fine-tune the backbone on the mixed dataset with the route-level $80/20$ split of Section~\ref{sec:framework:sft} of the main paper. Training uses $\mathcal{L}_{\text{SFT}}$ (Eq.~\ref{eq:sft_loss}) with prompts that contain $K = 2$ retrieved demonstrations. The backbone thus learns to read the demonstrations as few-shot examples and to imitate the driving they show, on both style-agnostic and style-specific data.
\end{itemize}
Hyperparameters and compute for all three stages are reported in
Sec.~\ref{app:hyperparameters}.

\paragraph{What does and does not change with style.}
Three things are shared across all styles. They are the SigLIP vision encoder $f_v$, the retrieval head $(f_c, f_{\text{ret}})$, and the fine-tuned backbone. Only one thing changes, which is the FAISS index searched in step~2. In principle, adding a fourth style would only require recording new style-instructed demonstrations, embedding them with the existing retrieval head, and building a new FAISS index. Whether the backbone generalizes to a style it was not fine-tuned on remains untested.

\section{Retrieval Head Architecture Details}
\label{app:retrieval_arch}

Table~\ref{tab:architecture_appendix} reports the full architecture
of the PersonaDrive retrieval head trained in Stage~2, including
input and output dimensions and the frozen-versus-trained status of
each component.

\begin{table*}[t]
\centering
\caption{Architecture of the PersonaDrive retrieval head (Stage~2).
\textbf{Frozen}: weights fixed throughout training.
\textbf{Trained}: weights updated by $\mathcal{L}_{\text{ret}}$.}
\label{tab:architecture_appendix}
\begin{tabular}{llccc}
\toprule
Component & Role & Input dim & Output dim & Status \\
\midrule
SigLIP $f_v$              & Vision encoder                       & $H{\times}W{\times}3$ & 768  & Frozen  \\
\addlinespace
LayerNorm         & Normalize scalars                    & 4                     & 4    & Trained \\
Scalar MLP        & Encode speed, throttle, steering, brake & 4                  & 32   & Trained \\
Command embedding         & Encode navigational command          & $|\mathcal{A}|$       & 16   & Trained \\
Fusion (Linear+GELU+LN)   & Merge scalar \& command              & 48                    & 128  & Trained \\
\addlinespace
Concat                    & Vision $\|$ control                  & $768 + 128$           & 896  & ---     \\
\addlinespace
ProjectionMLP $f_{\text{ret}}$ & Project to retrieval space      & 896                   & 1024 & Trained \\
L2-norm                   & Unit-normalize embedding             & 1024                  & 1024 & ---     \\
\bottomrule
\end{tabular}
\end{table*}

The ControlEncoder $f_c$ first normalizes the four control signals (speed, throttle, steering, and the binary brake flag) with a LayerNorm and passes them through a two-layer MLP with GELU activations and layer widths $4$, $32$, and $32$. In parallel, the navigational command is mapped to a learned $16$-dimensional embedding. The two branches are concatenated into a $48$-dimensional vector, which a linear layer followed by GELU and LayerNorm fuses into the $d_c = 128$-dimensional control vector. The ProjectionMLP $f_{\text{ret}}$, composed of two linear layers with GELU and LayerNorm, then maps the concatenated image and control embeddings ($768 + 128 = 896$ dimensions) to a $d_r = 1024$-dimensional output. This output is $L_2$-normalized for inner-product search with the FAISS \texttt{IndexFlatIP} index.

\begin{figure*}[t]
  \centering
  \includegraphics[width=\textwidth]{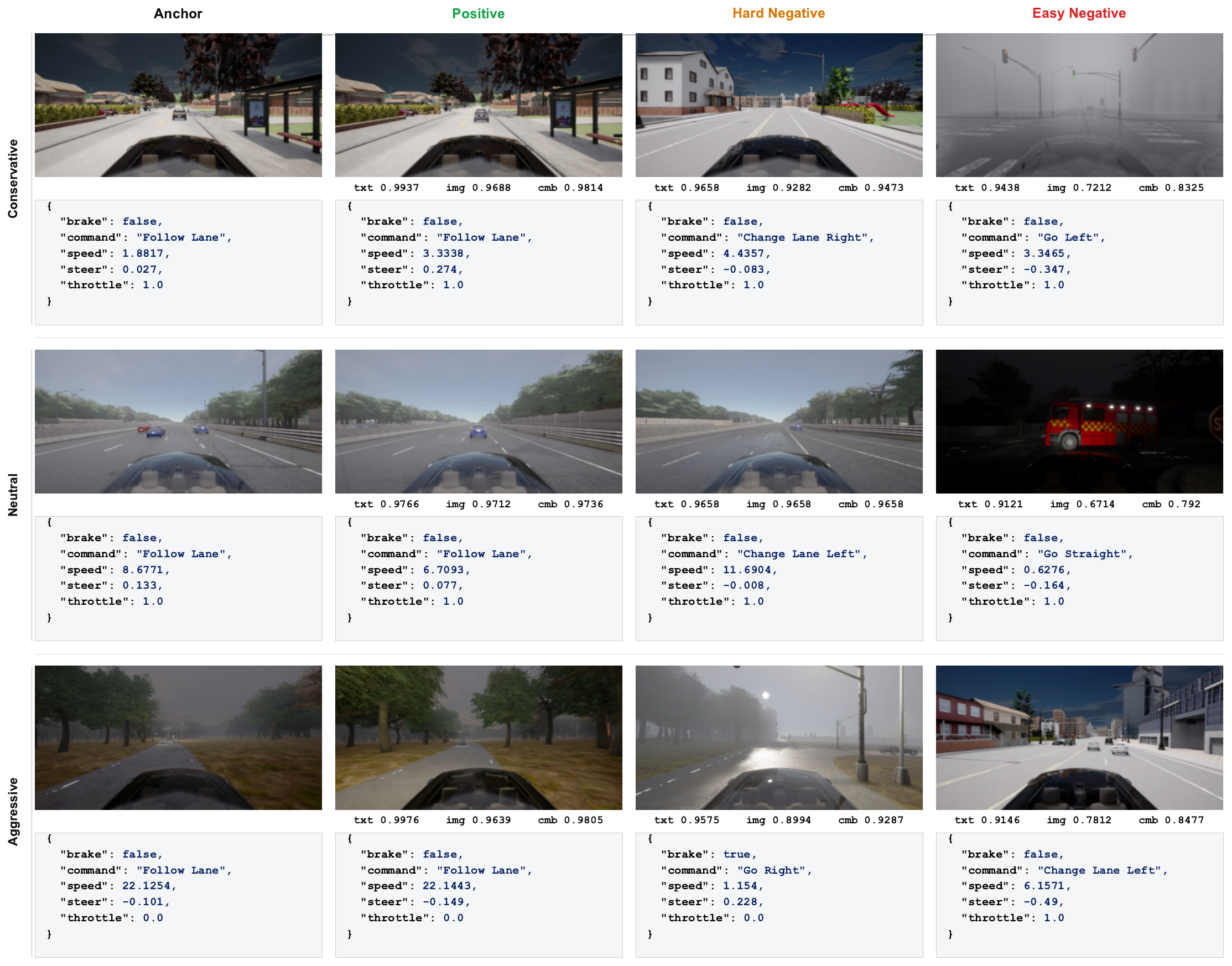}
  \caption{Examples of mined triplets for the three styles. Each row shows an anchor and its three partners. The positive has the highest combined similarity, with a nearly identical scene and the same command, so it calls for the same kind of driving. The hard negative has a high image similarity but a different command or speed. The easy negative has the lowest combined similarity and differs from the anchor in both scene and action.}
  \label{fig:triplet_viewer}
\end{figure*}

\section{Frame Selection: Failure-Mode Walkthrough}
\label{app:frame_selection}

This section explains why we keep only every fifth recorded frame of each route, for both the human and the PDM-lite data, before triplet mining (Section~\ref{sec:framework:triplets} of the main paper). Recorded frames are $0.25$\,s apart, so the kept frames are $1.25$\,s apart.

\paragraph{The trivial-positives failure mode.}
Two consecutive recorded frames differ by very little motion, so their scene, control signals, and resulting waypoints are almost the same. An anchor at frame $\tau$ and a candidate at frame $\tau{+}1$ would therefore have a similarity $\sigma^p_{\tau,\tau+1}$ close to $1$, because they are one step apart in time, not because they show similar driving. Without subsampling, such trivial positives would dominate the candidate pool and drown out the hard negatives, which look similar but call for different driving. The distances to positives and to hard negatives would then become nearly equal, and the training signal would collapse.

\paragraph{Why stride-of-5 specifically.}
A stride of $5$ leaves enough time between kept frames ($1.25$\,s) for the scene, the ego state, and the controls to change in a meaningful way, yet it is short enough that consecutive entries of one maneuver stay related. Strides below $3$ (under $0.75$\,s) led to the collapse described above, and strides above $8$ (over $2$\,s) began to skip key moments within a maneuver, such as the start of braking before a turn. We did not run a full ablation over the stride, so this choice remains a design decision based on these observations.

\paragraph{Anchor promotion.}
The kept frames form the candidate pool for triplet mining. A frame becomes an \emph{anchor} only if at least one valid easy negative and one valid hard negative exist for it under the combined score $\sigma^p_{\tau,\tau'}$. Frames without either kind of negative are left out of training. However, they stay in the database, because they can still be retrieved as nearest neighbors of query frames at inference.

\section{Style Data Collection}
\label{app:data_collection}

\begin{figure}[t]
  \centering
  \includegraphics[width=\linewidth]{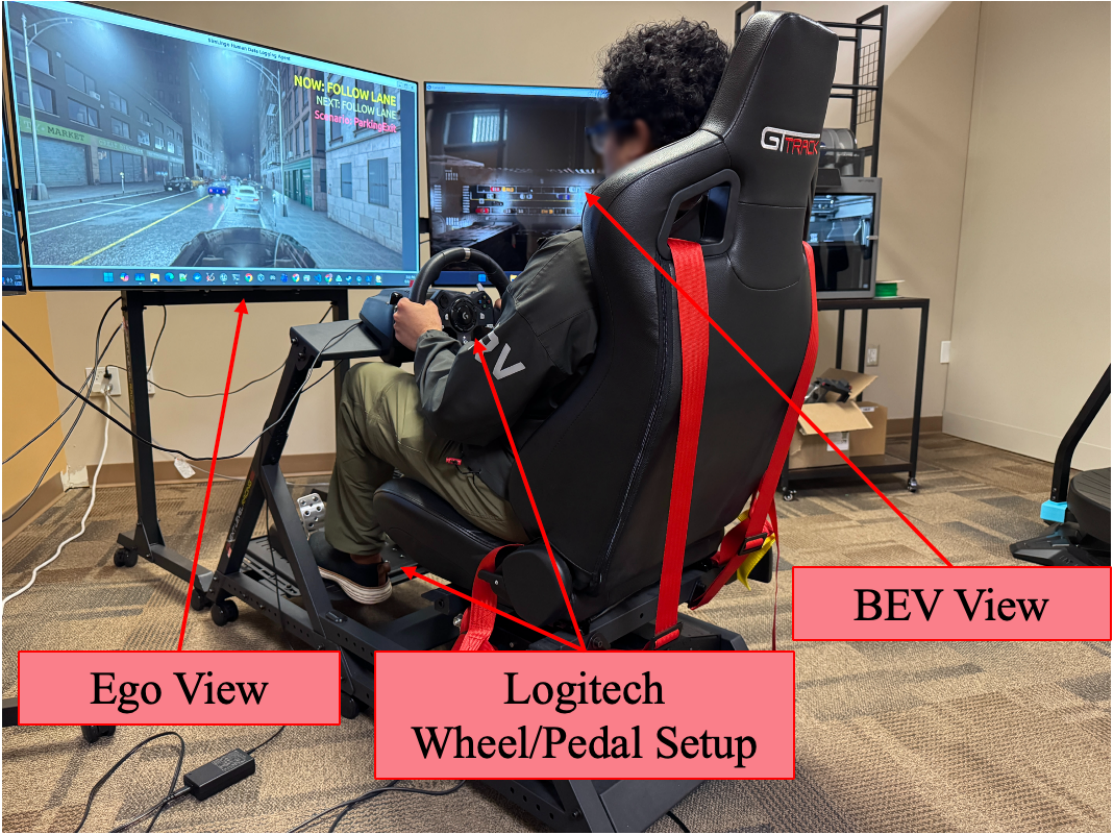}
  \caption{\textbf{Hardware setup used for data collection.} The Logitech wheel and pedals record the driver's control inputs, and the screens show the simulated scene.}
  \label{fig:hardware_rig}
\end{figure}

\paragraph{Hardware rig.}
We collected the style data offline from $M = 15$ participants, with a human driver in the loop. Each participant drives an ego vehicle in CARLA on a hardware rig (Fig.~\ref{fig:hardware_rig}). The rig has a Next Level Racing GTrack bucket seat, three 50-inch screens that give a field of view of about $130^\circ$, and a Logitech G923 steering wheel with pedals. The simulator runs in synchronous mode at $20$\,Hz. The ego vehicle is the only actor controlled by a human, and CARLA's TrafficManager controls the surrounding traffic. The ego view shows what is directly in front of the car, including the front-left and front-right views. A bird's-eye view (BEV) shows the surrounding vehicles, which helps with harder maneuvers such as merging into traffic.

\begin{figure*}[t]
  \centering
  \includegraphics[width=\textwidth]{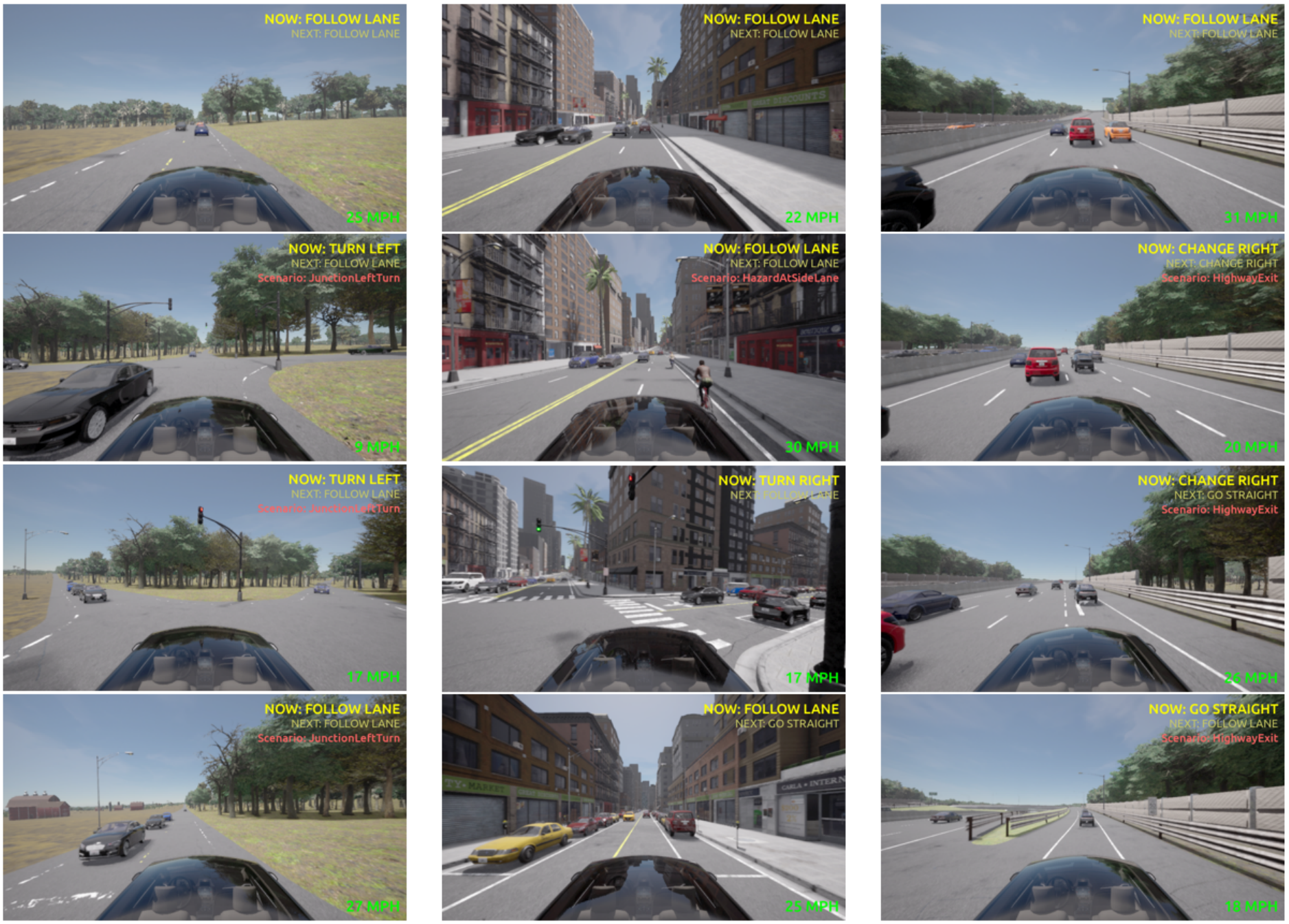}
  \caption{\textbf{Human data-collection interface.} Three examples of
  routes driven through the Pygame interface. The left column shows an
  unprotected left turn, the middle a hazard at the side of the lane, and
  the right a highway exit scenario.}
  \label{fig:datacollection_interface}
\end{figure*}

\paragraph{User interface.}
Participants drive CARLA through a Pygame window that shows the first-person view of the car, its speed, the current navigation instruction, and the active scenario. Fig.~\ref{fig:datacollection_interface} shows this interface in three scenarios. The left column is an unprotected left turn, where the second and third images show the navigation instruction telling the driver to turn left through the intersection. The middle column has a hazard at the side of the road. Once the hazard is in sight, a red scenario flag signals that the driver must handle it properly. The right column is a highway scenario in which the driver must move two lanes to the right to exit. Its second and third images show the navigation instruction for the lane change, and the fourth shows the end of the scenario after the exit.

\paragraph{Style instructions.}
Each participant drives each of their routes three times
(Figure~\ref{fig:data_collection}), once under each of three explicit style
instructions inspired by StyleDrive~\cite{hao2026styledrive}:
\begin{itemize}
    \item \textbf{Conservative.} Target $10$--$15$\,mph on surface streets and $25$--$35$\,mph on the highway, and yield to other vehicles whenever possible.
    \item \textbf{Neutral.} Follow traffic laws, navigate safely,
          and drive at a reasonable speed.
    \item \textbf{Aggressive.} Drive as fast as possible without crashing, and behave aggressively, for example by tailgating and by cutting off merging vehicles.
\end{itemize}
We repeat each route three times because a driver who knows the route can act more decisively under each instruction. The repetition also reflects how one person switches between styles in daily driving, for example driving conservatively on unfamiliar roads and aggressively when in a hurry. We also supervise the participants throughout the trials to make sure they follow the style instructions.

\paragraph{Dataset verification.}
We draw the routes at random from the CARLA Leaderboard~2.0 route set. Together, the participants' routes cover $21$ distinct Leaderboard~2.0 scenario types. The collection routes are disjoint from the $220$ Bench2Drive evaluation routes. Their scenario types also appear in Bench2Drive by design, so a retrieved demonstration can be relevant to an evaluation situation. However, no Bench2Drive evaluation episode is in any style database. To check that the participants followed the instructions, we compute speed statistics per frame from the recorded ego state. We use the $55$ routes driven under all three instructions (Table~\ref{tab:style_speed}). Both the overall mean speed and the mean speed while moving rise steadily from conservative to aggressive. The conservative mean while moving also falls inside the $10$--$15$\,mph target for surface streets. While moving, the aggressive drives are $90\%$ faster than the conservative ones. This gap is large even though most routes are urban, with dense traffic and frequent stops that limit the possible speed.

\begin{table}[t]
\centering
\caption{Speed statistics of the style-instructed human driving data, on the $55$ routes driven under all three instructions. \emph{In motion} leaves out frames slower than $0.2$\,mph ($0.1$\,m/s), and \emph{Stopped} is the share of such frames. Speeds are in mph.}
\label{tab:style_speed}
\small
\setlength{\tabcolsep}{3pt}
\begin{adjustbox}{max width=\linewidth}
\begin{tabular}{lcccc}
\toprule
\textbf{Instruction} & \textbf{Mean} & \textbf{In motion} & \textbf{Std.\ dev.} & \textbf{Stopped (\%)} \\
\midrule
Conservative & $9.8$  & $11.5$ & $7.6$  & $14.6$ \\
Neutral      & $13.2$ & $15.6$ & $10.8$ & $15.7$ \\
Aggressive   & $17.5$ & $21.8$ & $16.8$ & $19.8$ \\
\bottomrule
\end{tabular}
\end{adjustbox}
\end{table}

\paragraph{Participant consent.}
All participants gave informed consent before data collection. They were told the purpose of the study, the data being recorded (simulator state and control inputs, with no audio or video of the participant), and their right to withdraw. Only simulator data is recorded, and the released dataset contains no information that identifies a participant. The participants were non-professional drivers with valid driving licenses.

\paragraph{Recording and annotation.}
Every fifth simulator step ($4$\,Hz), we save the observation tuple $o^p_\tau$ together with the sensor and ground-truth streams. The ego camera gives the front-view RGB frame $\mathbf{I}^p_\tau$. The ego speed $v^p_\tau$ comes from CARLA's vehicle state, and the steering angle $\delta^p_\tau$ is read from the wheel. Throttle and brake are recorded from the pedals. CARLA's route planner provides the navigational command $c^p_\tau$ and the next GPS target points $\mathbf{g}^p_\tau$. We also compute a list of nearby objects, such as other vehicles, pedestrians, traffic signs, and traffic-light states. We use this list afterwards to generate visual question-answering annotations that follow the DriveLM schema~\cite{sima2024drivelm}, and free-text commentary that follows SimLingo~\cite{renz2025simlingo}. Both annotations are split by style into the three FAISS indices used at inference (Section~\ref{sec:framework:retrieval} of the main paper).

\paragraph{Agent behavior divergence.}

\begin{figure}[t]
  \centering
  \includegraphics[width=\linewidth]{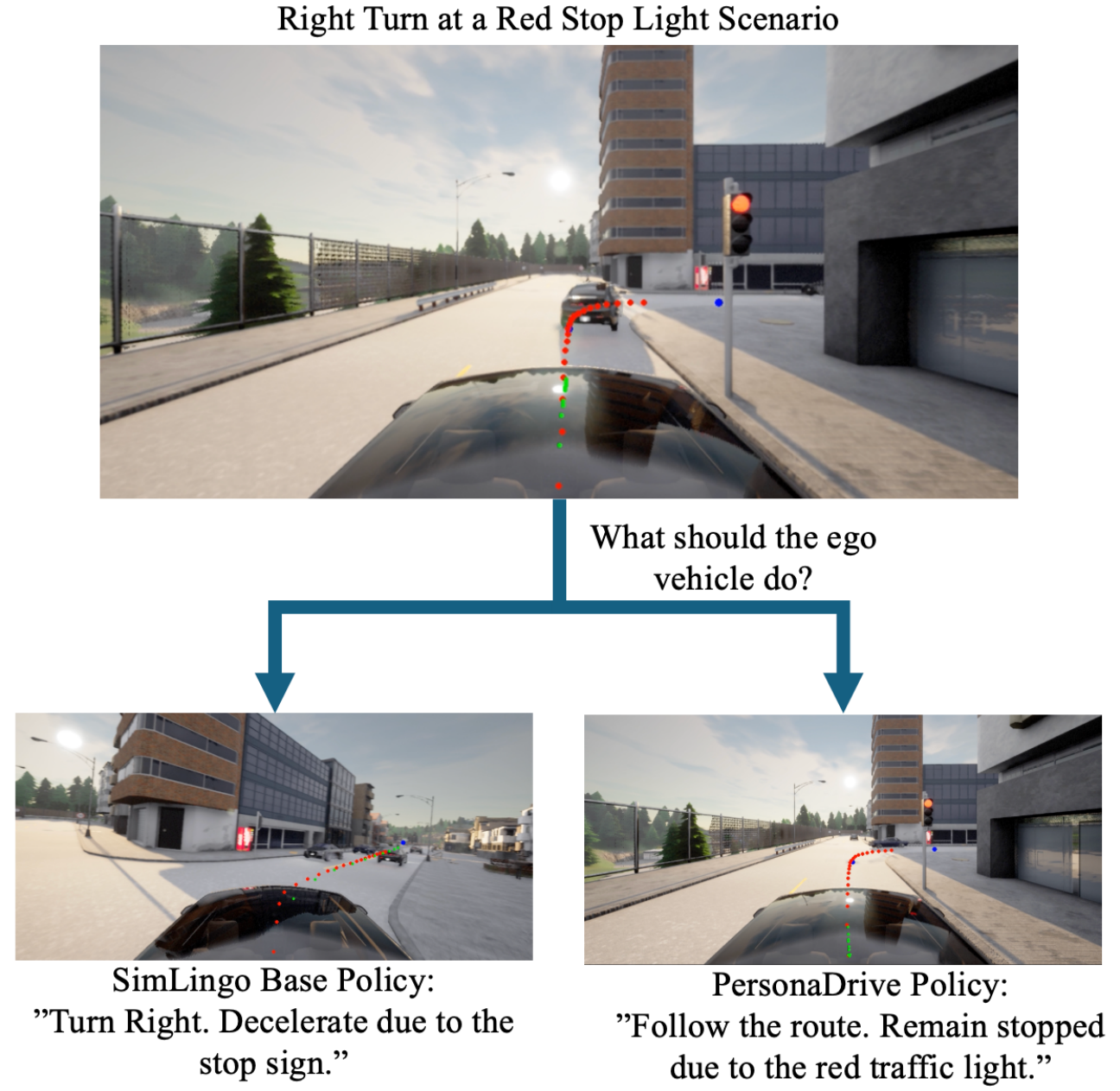}
  \caption{\textbf{Agent behavior divergence.} In the same scenario, the SimLingo base policy and PersonaDrive behave differently.}
  \label{fig:agent_behavior}
\end{figure}

In the closed-loop Bench2Drive evaluation, PersonaDrive does not simply copy the base SimLingo policy. It also shows new behaviors learned from the human demonstrations. Fig.~\ref{fig:agent_behavior} shows a scenario with a right turn at a red light. The SimLingo policy consistently drives through the red light and completes the turn with some slowing down. In contrast, PersonaDrive stays stopped and names the red light as the reason. This example shows that PersonaDrive behaves differently from the original SimLingo policy. A side-by-side video is on the project page (\url{https://pervasiveautonomylab.github.io/personadrive-landpage/}).

\paragraph{Style-conditioned behavior divergence.}
Table~\ref{tab:bench2drive} of the main paper (Section~\ref{sec:exp:b2d_style}) reports averages over all routes. To complement it, we show the same PersonaDrive checkpoint driving the same Bench2Drive route twice. The first run searches the aggressive index, and the second searches the conservative index. Nothing else differs between the two runs. Both runs meet the same cyclist ahead in the ego lane. The aggressive agent overtakes at once. It moves into the oncoming lane and accelerates past the cyclist at almost $19$\,m/s. The conservative agent instead stays behind the cyclist at about $1$--$3$\,m/s. It waits until the lane ahead is clear before it returns to its target speed. The two runs therefore finish the segment far apart in time. The aggressive agent clears it much earlier, while the conservative agent gives up progress to keep its distance from the cyclist. A side-by-side video is on the project page (\url{https://pervasiveautonomylab.github.io/personadrive-landpage/}).

\section{Training Data Sampling and Scenario Coverage}
\label{app:training_data}

PersonaDrive uses two separate subsets of SimLingo's PDM-lite training data~\cite{renz2025simlingo}, one for each trained component. A $5\%$ subset trains the retrieval head, and a $20\%$ subset of the PDM-lite data is used for the in-context SFT of the backbone. Both subsets are sampled by \emph{route}. Each unit is a complete PDM-lite route, not a random set of frames from the whole collection of $3.1$ million samples.

\paragraph{Why route-level sampling.}
Sampling whole routes keeps each drive coherent, with the natural sequence of merging, cruising, turning, and obstacle handling that PDM-lite drives along a long Leaderboard~2.0 route. The retrieval system needs this context within an episode to learn from, and the backbone needs it at inference. Sampling random frames would break this structure over time.

\paragraph{Why 5\% versus 20\%.}
The trained part of the retrieval head is small, the ControlEncoder plus the ProjectionMLP, and it is trained with a contrastive loss. A few hundred thousand triplets are enough to learn an embedding that reflects driving behavior. Teaching the backbone to read the in-context prompt (Eq.~\ref{eq:prompt} of the main paper) and to produce waypoints that fit the style needs much more supervision. The SFT slice is therefore much larger.

\paragraph{Scenario coverage.}
We deliberately choose the \emph{PDM-lite} routes of Experiment~1, not the human routes, to cover the scenario types of the CARLA Leaderboard~\cite{carla_lb1_scenarios}. These include \emph{Control Loss}, \emph{Unprotected Left Turn}, \emph{Right Turn with Crossing Traffic}, \emph{Crossing Negotiation}, \emph{Highway Merge}, \emph{Static Cut-In}, \emph{Obstacle in Lane}, \emph{Pedestrian Emerging from Behind Parked Vehicle}, and \emph{Parking Cut-in}, among others. The retrieval database therefore contains demonstrations of each type. Both the $5\%$ and the $20\%$ slice thus span many different scenario types rather than a single kind of situation.

\section{Hyperparameters}
\label{app:hyperparameters}

Table~\ref{tab:hyperparameters} lists all hyperparameters used in
the three training stages.

\begin{table}[t]
\centering
\caption{Hyperparameters for the three PersonaDrive training stages.
$N^{+}$ and $N^{-}$ are the top / bottom percentile
thresholds bounding the Stage~1 positive / easy-negative pools.}
\label{tab:hyperparameters}
\small
\setlength{\tabcolsep}{4pt}
\begin{adjustbox}{max width=\linewidth}
\begin{tabular}{lll}
\toprule
\textbf{Hyperparameter} & \textbf{Symbol} & \textbf{Value} \\
\midrule
\multicolumn{3}{l}{\textit{Stage 1: Triplet mining}} \\
Image similarity weight                 & $\lambda_{\text{img}}$ & $0.5$ \\
Text similarity weight                  & $\lambda_{\text{txt}}$ & $0.5$ \\
Positive pool threshold (top pct.)      & $N^{+}$ & $5\%$ \\
Easy-negative threshold (bottom pct.)   & $N^{-}$ & $5\%$ \\
Stride (frame subsampling)              & ---                    & $5$ \\
\midrule
\multicolumn{3}{l}{\textit{Stage 2: Retrieval head training}} \\
Triplet loss margin                     & $\beta$ & $0.3$ \\
Hard-negative weight                    & $w_h$   & $2.0$ \\
Vision dim                              & $d_v$   & $768$ \\
Control dim                             & $d_c$   & $128$ \\
Retrieval dim                           & $d_r$   & $1024$ \\
Optimizer                               & ---     & AdamW \\
Learning rate                           & ---     & $1\times 10^{-4}$ \\
Batch size                              & ---     & $128$ triplets \\
Epochs                                  & ---     & $20$ \\
\midrule
\multicolumn{3}{l}{\textit{Stage 3: Supervised fine-tuning of VLA backbone}} \\
Top-$K$ retrieved demonstrations        & $K$        & $2$ \\
Pos/vel loss balance                    & $\alpha$   & $1.0$ \\
Language-loss weight                    & $\lambda_{\text{lang}}$ & $1.0$ \\
Waypoints per demonstration     & $N$        & $10$ \\
Optimizer                               & ---        & AdamW \\
Learning rate                   & ---        & $3\times 10^{-5}$ \\
Batch size                      & ---        & $24$ prompts ($6$ per GPU $\times$ $4$ GPUs) \\
Epochs                          & ---        & $1$ \\
LoRA~\cite{hu2022lora} rank (whole language model) & ---       & $32$ \\
LoRA scaling factor             & ---        & $64$ \\
Also trained                    & ---        & projector, driving adaptor, waypoint encoder \\
\bottomrule
\end{tabular}
\end{adjustbox}
\end{table}

\paragraph{Compute.}
We ran all retrieval-head training and all backbone SFT on four NVIDIA L40S GPUs ($48$\,GB each), using DeepSpeed ZeRO-2 to split memory. A full Stage~3 SFT run takes about $12$ hours on these four GPUs. Bench2Drive closed-loop evaluation runs on a single L40S and takes about $12$ hours per style.

\paragraph{Notes on hyperparameter selection.}
We chose the triplet margin $\beta$ and the hard-negative weight $w_h$ from a small grid ($\beta \in \{0.1, 0.2, 0.3, 0.5\}$ and $w_h \in \{1.5, 2.0, 3.0\}$). We used a held-out validation slice of the Stage~1 mining data and kept the combination with the largest gap between hard-negative and positive distances at the end of training. All other hyperparameters come from SimLingo's defaults or are standard values for the optimizers used, and we did not tune them further.

\section{Context-Point Tuple Formalization}
\label{app:context_point}

This section gives the full formalization of the context point
$\xi^p_\tau$ summarized in Section~\ref{sec:framework:context_point}
of the main paper.

\paragraph{Formal tuple definition.}
Each entry in the style-$p$ database is a context point
$\xi^p_\tau$, defined as the six-field tuple
\begin{equation}
\label{eq:context_point}
\begin{aligned}
\xi^p_\tau \;=\; \Big(\,
    & \underbrace{\mathbf{I}^p_\tau}_{\text{RGB frame}},\
      \underbrace{\mathbf{Q}^p_{\tau-2:\tau}}_{\text{control history}},\
      \underbrace{c^p_\tau}_{\text{command}}, \\[6pt]
    & \underbrace{\mathbf{g}^p_\tau}_{\text{2 target points}},\
      \underbrace{r^p_\tau}_{\text{commentary}},\
      \underbrace{\mathbf{W}^{p,\star}_\tau \in \mathbb{R}^{N \times 2}}_{\text{future waypoints}} \,\Big).
\end{aligned}
\end{equation}
Here $c^p_\tau \in \mathcal{A}$ is the navigational command at frame $\tau$. The matrix $\mathbf{g}^p_\tau \in \mathbb{R}^{2 \times 2}$ holds the next two GPS target points in the ego frame (target\_point and target\_point\_next, as in SimLingo). The field $r^p_\tau$ is the commentary, which describes the reasoning behind the action. Finally, $\mathbf{W}^{p,\star}_\tau \in \mathbb{R}^{N \times 2}$ are the $N = 10$ waypoints driven under style $p$ after frame $\tau$. There is one waypoint per recorded frame at $4$\,Hz, so they cover the next $2.5$\,s. The VQA annotation $q^p_\tau$ is not one of the six fields. It is stored alongside the context point as an extra label, which only Stage~1 mining uses.

\paragraph{Control history matrix.}
The field $\mathbf{Q}^p_{\tau-2:\tau}$ is the control history of the anchor frame and the two frames before it. It records two signals, speed and throttle, at each of the three steps:
\begin{equation}
\label{eq:control_history}
\setlength{\arraycolsep}{2pt}
\mathbf{Q}^p_{\tau-2:\tau} = \begin{pmatrix}
v^p_{\tau-2}              & v^p_{\tau-1}              & v^p_{\tau}              \\[2pt]
\text{throttle}^p_{\tau-2} & \text{throttle}^p_{\tau-1} & \text{throttle}^p_{\tau}
\end{pmatrix} \in \mathbb{R}^{2 \times 3},
\end{equation}
where each row is one control signal and each column is one time step, from oldest to newest. This $2 \times 3$ history tells the backbone how the situation developed before the demonstrated action $\mathbf{W}^{p,\star}_\tau$. For example, it shows whether the driver was slowing down, speeding up, or cruising at a steady speed before frame $\tau$.

\paragraph{Why no separate ego-state field.}
The current speed and throttle are already the last column of $\mathbf{Q}^p_{\tau-2:\tau}$, so a separate ego-state field would only repeat them. The steering angle and the brake flag are not stored in the context point, but they are still used elsewhere. The retrieval head takes both as input (Sec.~\ref{app:retrieval_arch}), and the Stage~1 frame sentence includes the steering angle (Eq.~\ref{eq:sentence}).

\paragraph{Brake.}
The binary brake flag is an input to the retrieval head's
ControlEncoder (Sec.~\ref{app:retrieval_arch}) but is not part of
the control history $\mathbf{Q}^p_{\tau-2:\tau}$.

\paragraph{Database compactness.}
Storing the control history inside each context point keeps the database small. The backbone still gets a compact summary of the recent motion, while the database stores only six extra numbers per entry instead of two extra frames with images.

\paragraph{Three roles of the tuple fields.}
The fields of a context point, together with the VQA label stored alongside it, play three different roles.
\begin{itemize}
    \item \textbf{Retrieval} (Eq.~\ref{eq:retrieval_embedding}).
          The retrieval embedding $\mathbf{s}^p_\tau$ is computed from image and control signals only. It is used only for the nearest-neighbor search in the FAISS index, and no text or waypoint field takes part in retrieval at inference.
    \item \textbf{Mining} (Eq.~\ref{eq:combined_sim}).
          The VQA annotation $q^p_\tau$ is used only in Stage~1 triplet mining. There, together with the commentary $r^p_\tau$ and the other fields, it forms the sentence $e^p_\tau$ that drives the combined similarity score $\sigma^p_{\tau,\tau'}$. It is stored alongside the context point rather than as one of its six fields, and it is never written into the prompt of the backbone.
    \item \textbf{Serialization} (Eq.~\ref{eq:serialize}).
          All six fields of the context point form an in-context demonstration during SFT and inference. They are the image, the control history, the command, the two target points, the commentary $r^p_\tau$, and the driven waypoints. The waypoints $\mathbf{W}^{p,\star}_\tau$ carry the most information about behavior. They show how the human actually responded to the situation under style $p$, through the shape of the path and through the speed, which follows from the spacing between waypoints.
\end{itemize}

\section{Retrieval Depth Ablation: Full Cost Breakdown}
\label{app:k-depth}

This section extends Table~\ref{tab:k-depth-main} of the main paper with the full cost of each step. The setup is the same. We vary $K$ from $0$ to $4$ in the no-style setting of Section~\ref{sec:exp:b2d_base}. $K = 0$ is the SimLingo backbone without retrieval and matches the SimLingo row of Table~\ref{tab:b2d_base}. For $K \geq 1$, we use the PersonaDrive backbone with $K$ demonstrations retrieved from the style-agnostic PDM-lite index of Section~\ref{sec:exp:b2d_base}. The $K = 2$ column matches the PersonaDrive (no style) row of Table~\ref{tab:b2d_base}. For $K \geq 1$, the ablation therefore changes only the number of demonstrations, while $K = 0$ is a reference point that uses a different backbone (the original SimLingo). We measure the open-loop errors on held-out PDM-lite routes that are not in the $5\%$ and $20\%$ training slices. ADE is the mean distance between the predicted and the recorded waypoints over the $10$ temporal waypoints ($2.5$\,s), and FDE is the distance at the last waypoint. Table~\ref{tab:k-depth} gives the full sweep, including the time rows left out of Table~\ref{tab:k-depth-main}.

\begin{table}[t]
\centering
\caption{\textbf{Retrieval-depth ablation in the no-style setting} ($K$ from $0$ to $4$). $K = 0$ is the SimLingo backbone, and the $K = 2$ column matches the PersonaDrive (no style) row of Table~\ref{tab:b2d_base}. ADE and FDE are open-loop errors in meters on held-out PDM-lite routes. DS, SR, Eff., and Comf.\ are Bench2Drive closed-loop metrics. The time rows are averages per step in milliseconds. TTFT is the time to process the prompt (prefill), and TTLT is the time to generate the output (decode). Inference total also includes the dual waypoint head and the assembly of each demonstration, which runs the backbone's vision encoder on every retrieved image. This part therefore grows with $K$. RAG overhead is the FAISS search time, so Total equals Inference total plus RAG overhead. Higher Comfort means smoother driving, as in Bench2Drive.}
\label{tab:k-depth}
\small
\setlength{\tabcolsep}{3pt}
\begin{adjustbox}{max width=\linewidth}
\begin{tabular}{lcc>{\columncolor{oursrow}}ccc}
\toprule
\textbf{Metric} & $K = 0$ & $K = 1$ & $K = 2$ & $K = 3$ & $K = 4$ \\
\midrule
\multicolumn{6}{l}{\textit{Cost (per-step averages)}} \\
RAG overhead (ms)        & 0.0   & 19.45 & 19.59 & 19.60 & 19.60 \\
Tokens                   & 547   & 1292  & 2030  & 2769  & 3507  \\
TTFT (ms)\,$\downarrow$   & 34.0  & 45.1  & 70.3  & 95.9  & 125.3 \\
TTLT (ms)\,$\downarrow$   & 424.1 & 425.6 & 427.3 & 427.6 & 430.0 \\
Inference total (ms)\,$\downarrow$ & 480.3 & 490.1 & 538.1 & 579.8 & 630.1 \\

Total (ms)\,$\downarrow$  & 480.3 & 509.6 & 557.7 & 599.4 & 649.7 \\
\midrule
\multicolumn{6}{l}{\textit{Open-loop accuracy (held-out PDM-lite routes)}} \\
ADE (m)\,$\downarrow$ & 0.820 & 0.880 & 0.795 & 0.793 & 0.791 \\
FDE (m)\,$\downarrow$ & 1.787 & 1.961 & 1.699 & 1.650 & 1.648 \\
\midrule
\multicolumn{6}{l}{\textit{Closed-loop (Bench2Drive)}} \\
DS\,$\uparrow$        & 85.07 & 81.55 & 88.95 & 89.16 & 89.30 \\
SR\,(\%)\,$\uparrow$  & 67.27 & 62.17 & 72.29 & 72.35 & 73.10 \\
Eff.\,$\uparrow$      & 259.23 & 259.45& 255.15& 261.82& 261.01 \\
Comf.\,$\uparrow$     & 33.67 & 25.77 & 28.09 & 30.24 & 33.80 \\
\bottomrule
\end{tabular}
\end{adjustbox}
\end{table}

\paragraph{Findings.}
Beyond $K = 2$, the two metrics we optimize barely change. Going from $K = 2$ to $K = 4$ raises the Driving Score by only $0.4\%$ and the Success Rate by $1.1\%$, which is within the variation between runs of the benchmark. However, the cost of these small gains grows linearly with $K$. Each added demonstration adds about $740$ prompt tokens. Going from $K = 2$ to $K = 4$ therefore makes the prompt $73\%$ longer, the prefill time $78\%$ longer, and the total time per step $16.5\%$ longer. Compared with the SimLingo backbone at $K = 0$, $K = 2$ adds $16.1\%$ to the time per step. Only a quarter of this extra time is the FAISS search, and the rest comes from the longer in-context block. The decode time (TTLT) and the FAISS time stay almost constant for every $K \geq 1$. The main slowdown therefore comes from processing the longer in-context block, not from retrieval itself. This time is the main constraint in our setting. Closed-loop driving pays it at every step, and running several style-conditioned agents in one simulation, which we leave to future work, would multiply it. We therefore use $K = 2$, the smallest depth that gives the full gain in Driving Score and Success Rate. Efficiency and Comfort are $2.3\%$ and $20\%$ higher at $K = 4$ than at $K = 2$. Our choice is therefore a deliberate trade of some smoothness for speed, not a claim that no metric improves. At the other end, a single demonstration ($K = 1$) is not enough, and its Driving Score is $4.1\%$ below the baseline without retrieval.
\section{Supervised Fine-Tuning Loss}
\label{app:sft_loss}

This section gives the full Stage~3 fine-tuning loss summarized in Section~\ref{sec:framework:sft} of the main paper. We keep SimLingo's objective~\cite{renz2025simlingo} unchanged. It supervises both outputs of the backbone together, the waypoints by regression and the language by next-token prediction:
\begin{equation}
\label{eq:sft_loss}
\mathcal{L}^p_{\text{SFT}} \;=\; \mathcal{L}^p_{\text{wp}} \;+\; \lambda_{\text{lang}} \, \mathcal{L}^p_{\text{CE}},
\end{equation}
where the waypoint loss is
\begin{equation}
\label{eq:sft_loss_wp}
\begin{aligned}
\mathcal{L}^p_{\text{wp}} \;=\;
    & \displaystyle\frac{1}{20} \sum_{n=1}^{20} \mathrm{SL}_1\!\big( \mathbf{W}^{\text{pos},n}_t,\, \mathbf{W}^{p,\star,\text{pos},n}_t \big) \\
  + \; & \alpha \, \displaystyle\frac{1}{10} \sum_{n=1}^{10} \mathrm{SL}_1\!\big( \mathbf{W}^{\text{vel},n}_t,\, \mathbf{W}^{p,\star,\text{vel},n}_t \big).
\end{aligned}
\end{equation}
Here $\mathrm{SL}_1(\cdot,\cdot)$ is the smooth-$L_1$ loss summed over the two coordinates. The terms $\mathbf{W}^{\text{pos},n}_t$ and $\mathbf{W}^{\text{vel},n}_t$ are the $n$-th predicted geometric and temporal waypoints. The terms $\mathbf{W}^{p,\star,\text{pos},n}_t$ and $\mathbf{W}^{p,\star,\text{vel},n}_t$ are the matching ground-truth values from the style-$p$ trajectory. The term $\mathcal{L}^p_{\text{CE}}$ is the cross-entropy over the generated language tokens, and $\alpha$ and $\lambda_{\text{lang}}$ balance the three terms. As in SimLingo~\cite{renz2025simlingo}, we use the smooth-$L_1$ loss, which trains more stably than $L_2$, and equal weights $\alpha = \lambda_{\text{lang}} = 1$. The total SFT loss $\mathcal{L}_{\text{SFT}}$ averages $\mathcal{L}^p_{\text{SFT}}$ over all training examples, from the three styles and the style-agnostic PDM-lite data. Sec.~\ref{app:hyperparameters} reports the hyperparameter values.

\newpage
\section{Inference-Time Prompt Example}
\label{app:prompt_example}

This section gives a worked example of the in-context prompt $\mathcal{X}_t$ of Eq.~\ref{eq:prompt} in the main paper. It has $K = 2$ retrieved demonstrations followed by the live query observation, and it follows the field order of Eq.~\ref{eq:serialize}. Images and coordinate lists are shown by placeholder names. In the real prompt, images become tokens of the vision encoder, and coordinates are written as text (Section~\ref{sec:framework:sft} of the main paper).

\begin{quote}
\small
\texttt{<DEMO 1>} \\
\texttt{<image>} $\mathbf{I}^p_{(1)}$ \texttt{</image>} \\
\texttt{speed\_history: [4.21, 4.45, 4.62] m/s} \\
\texttt{throttle\_history: [0.62, 0.71, 0.78]} \\
\texttt{command: Follow Lane} \\
\texttt{target\_points: <coords> $\mathbf{g}^p_{(1)}$ </coords>} \\
\texttt{commentary: ``Maintaining cruise behind a slower lead vehicle.''} \\
\texttt{waypoints: <coords> $\mathbf{W}^{p,\star}_{(1)}$ </coords>} \\[3pt]
\texttt{<DEMO 2>} \\
\texttt{<image>} $\mathbf{I}^p_{(2)}$ \texttt{</image>} \\
\texttt{speed\_history: [3.95, 4.10, 4.30] m/s} \\
\texttt{throttle\_history: [0.55, 0.63, 0.68]} \\
\texttt{command: Follow Lane} \\
\texttt{target\_points: <coords> $\mathbf{g}^p_{(2)}$ </coords>} \\
\texttt{commentary: ``Slight rightward lane bias on a gentle curve.''} \\
\texttt{waypoints: <coords> $\mathbf{W}^{p,\star}_{(2)}$ </coords>} \\[3pt]
\texttt{<QUERY>} \\
\texttt{<image>} $\mathbf{I}_t$ \texttt{</image>} \\
\texttt{speed: 4.40 m/s} \\
\texttt{command: Follow Lane} \\
\texttt{target\_points: <coords> $\mathbf{g}_t$ </coords>} \\[3pt]
\texttt{<QUESTION> What should the ego do next?} \\
\texttt{<ANSWER> commentary: ``Following the lead vehicle at a safe distance.''} \\
\texttt{<QUERY TOKENS>} ($20 + 10$), decoded by the Route MLP into $\hat{\mathbf{W}}_t$
\end{quote}

The numerical values shown above are illustrative. Each
demonstration's speed and throttle history, command, target points,
commentary, and waypoints are serialized as text, and its image is
encoded by the backbone's InternViT vision encoder. In Stage~3, the backbone learns to write the commentary after \texttt{<ANSWER>}. The Route MLP then produces $\hat{\mathbf{W}}_t$ from the hidden states at the query tokens (Section~\ref{sec:framework:sft} of the main paper).

\onecolumn
\section{Notation Glossary}
\label{app:notation}

Table~\ref{tab:notation} consolidates the symbols used in the
main paper and this appendix.

{\small
\begin{longtable}{lp{12cm}}
\caption{Notation used throughout the main paper and appendix.}\label{tab:notation}\\
\toprule
\textbf{Symbol} & \textbf{Meaning} \\
\midrule
\endfirsthead
\caption[]{Notation used throughout the main paper and appendix (continued).}\\
\toprule
\textbf{Symbol} & \textbf{Meaning} \\
\midrule
\endhead
\bottomrule
\endfoot
\bottomrule
\endlastfoot
\multicolumn{2}{l}{\textit{Indices and sets}} \\*
$p \in \mathcal{P}$ & Style label, with $\mathcal{P} = \{\text{conservative}, \text{neutral}, \text{aggressive}\}$ \\
$\tau, t$           & Frame index in a database / simulator step of the live observation. Symbols without $p$, such as $\mathbf{I}_t$, $v_t$, $c_t$, and $\mathbf{g}_t$, refer to the live observation \\
$\mathcal{H}^p$     & Style-$p$ history (database of the stride-subsampled frames collected under style $p$) \\
$T^p$               & Number of context points in the style-$p$ database, $T^p = |\mathcal{H}^p|$ \\
$\mathcal{P}'$ & $\mathcal{P} \cup \{\text{style-agnostic}\}$, the three styles plus the style-agnostic PDM-lite slice used to train the retrieval head \\

$\mathcal{A}$       & Set of navigational commands (CARLA Leaderboard vocabulary: \textsc{follow lane}, \textsc{go left}, \textsc{go right}, \textsc{go straight}, \textsc{change lane left}, \textsc{change lane right}) \\
$M$                 & Number of human participants ($M = 15$) \\
\midrule
\multicolumn{2}{l}{\textit{Observations and tuple fields}} \\*
$o^p_\tau, o_t$        & Recorded observation at frame $\tau$ / live observation at step $t$ \\
$\mathbf{I}^p_\tau$ & Front-view RGB image at frame $\tau$ ($\mathbf{I}_t$ for the live observation) \\
$v^p_\tau, \delta^p_\tau$ & Ego speed and steering angle at frame $\tau$ \\
$c^p_\tau$          & Navigational command at frame $\tau$ \\
$\mathbf{g}^p_\tau \in \mathbb{R}^{2\times 2}$ & Two GPS target points in ego frame \\
$q^p_\tau, r^p_\tau$ & VQA annotation (mining-only label stored alongside $\xi^p_\tau$) and free-form commentary \\
$\mathbf{u}^p_\tau \in \mathbb{R}^{3}$ & Control scalars at frame $\tau$ (speed, throttle, steering) \\
$b^p_\tau \in \{0, 1\}$ & Brake flag at frame $\tau$ (retrieval-head input only) \\
$\mathbf{Q}^p_{\tau-2:\tau} \in \mathbb{R}^{2\times 3}$ & Control history (speed, throttle) at the anchor frame and the two frames before it \\
$\mathbf{W}^{p,\star}_\tau \in \mathbb{R}^{N\times 2}$ & The $N$ waypoints driven after frame $\tau$ under style $p$, i.e., the ego positions in the next $N$ recorded frames ($0.25$\,s apart) \\
$\xi^p_\tau$        & Context point: six-field tuple stored per database entry (Eq.~\ref{eq:context_point}) \\
\midrule
\multicolumn{2}{l}{\textit{Models and embeddings}} \\*
$f_v$               & Frozen SigLIP vision encoder \\
$f_t$               & Frozen BGE-M3 sentence encoder (Stage~1 mining only) \\
$f_c$               & Trained ControlEncoder, part of the retrieval head \\
$f_{\text{ret}}$    & Trained ProjectionMLP, part of the retrieval head \\
$\mathbf{s}^p_\tau, \mathbf{s}^q_t$ & Database / query retrieval embeddings ($\mathbb{R}^{d_r}$, $L_2$-normalized) \\
$d_v, d_c, d_r$     & Vision (768), control (128), retrieval (1024) embedding dimensions \\
$e^p_\tau$          & Seven-field frame sentence, used only for Stage~1 mining (Eq.~\ref{eq:sentence}) \\
\midrule
\multicolumn{2}{l}{\textit{Mining and similarity}} \\*
$\sigma^p_{\tau,\tau'}$ & Combined image-text similarity score (Eq.~\ref{eq:combined_sim}) \\
$\mathrm{sim}(\cdot,\cdot)$ & Cosine similarity \\
$\lambda_{\text{img}}, \lambda_{\text{txt}}$ & Mining-time image / text weights ($\lambda_{\text{img}}+\lambda_{\text{txt}}=1$) \\
$N^{+}, N^{-}$      & Top / bottom percentile thresholds bounding the positive / easy-negative pools \\
$\xi^{p,+}, \xi^{p,-}_{\text{easy}}, \xi^{p,-}_{\text{hard}}$ & Positive, easy-negative, and hard-negative samples, with $\xi^{p,-}$ for either kind of negative \\
$\mathbf{s}^{p,a}, \mathbf{s}^{p,+}, \mathbf{s}^{p,-}$ & Embeddings (Eq.~\ref{eq:retrieval_embedding}) of the anchor, positive, and negative frames \\
\midrule
\multicolumn{2}{l}{\textit{Losses and training}} \\*
$\beta$             & Triplet loss margin \\
$w_h$               & Hard-negative weight in Eq.~\ref{eq:weighted_triplet}, where $w = w_h$ for hard and $w = 1$ for easy negatives \\
$\mathcal{L}^p_{\text{tri}}, \mathcal{L}^p_{\text{ret}}, \mathcal{L}_{\text{ret}}$ & Triplet (Eq.~\ref{eq:base_triplet}), weighted triplet (Eq.~\ref{eq:weighted_triplet}), and total retrieval losses \\
$\mathcal{L}^p_{\text{SFT}}, \mathcal{L}_{\text{SFT}}$ & Per-style and averaged SFT losses (Eq.~\ref{eq:sft_loss}) \\
$\mathcal{L}^p_{\text{wp}}, \mathcal{L}^p_{\text{CE}}$ & Waypoint and language cross-entropy terms of the SFT loss (Eq.~\ref{eq:sft_loss_wp}) \\
$\alpha$            & Geometric / temporal waypoint SFT loss balance \\
$\lambda_{\text{lang}}$ & Weight on the language cross-entropy term of the SFT loss \\
\midrule
\multicolumn{2}{l}{\textit{Inference and prompt}} \\*
$K$                 & Number of retrieved demonstrations per query ($K = 2$) \\
$\mathcal{C}^p_t$   & Top-$K$ retrieved context points for the query at step $t$ under style $p$ (Eq.~\ref{eq:topk_retrieval}) \\
$\xi^p_{(k)}$ & The $k$-th retrieved context point in $\mathcal{C}^p_t$, serialized with Eq.~\ref{eq:serialize} \\
$\mathcal{X}_t$      & Serialized in-context prompt at step $t$ (Eq.~\ref{eq:prompt}) \\
$\hat{\mathbf{W}}_t = (\mathbf{W}^{\text{pos}}_t, \mathbf{W}^{\text{vel}}_t)$ & Predicted waypoint output ($20$ geometric, $10$ temporal) \\
$\mathbf{W}^{p,\star,\text{pos}}_t, \mathbf{W}^{p,\star,\text{vel}}_t$ & Ground-truth geometric and temporal waypoints of the current frame (Stage~3 targets, Eq.~\ref{eq:sft_loss_wp}) \\
$N$         & Waypoints per demonstration ($N = 10$) \\
\end{longtable}
}

\end{document}